\documentclass[conference]{IEEEtran}
\makeatletter
\def\ps@IEEEtitlepagestyle{%
  \def\@oddfoot{\mycopyrightnotice}%
  \def\@evenfoot{}%
}
\def\mycopyrightnotice{%
  {\footnotesize XXX-X-XXXX-XXXX-X/XX/\$XX.00~\copyright~20XX IEEE\hfill}% <--- Change here
  \gdef\mycopyrightnotice{}
}

\usepackage{blindtext}
\usepackage{tikz}
\usepackage{eso-pic}
\IEEEoverridecommandlockouts
\usepackage{float}
\usepackage{cite}
\usepackage{amsmath,amssymb,amsfonts}
\usepackage{algorithmicx}
\usepackage[ruled,vlined]{algorithm2e}
\usepackage{textcomp}
\usepackage{xcolor}
\usepackage{listings}
\usepackage{booktabs}
\def\BibTeX{{\rm B\kern-.05em{\sc i\kern-.025em b}\kern-.08em
    T\kern-.1667em\lower.7ex\hbox{E}\kern-.125emX}}
    
\usepackage{eso-pic}

\usepackage{graphicx,wrapfig}
\usepackage{textcomp}
\usepackage{xcolor}
\usepackage{algpseudocode}
\usepackage{caption}
\usepackage{subcaption}
\usepackage{comment}
\usepackage[font=scriptsize]{caption}
\usepackage{url}
\usepackage{txfonts}
\usepackage{array}
\definecolor{codegreen}{rgb}{0,0.6,0}
\definecolor{codegray}{rgb}{0.5,0.5,0.5}
\definecolor{codepurple}{rgb}{0.58,0,0.82}
\definecolor{backcolour}{rgb}{0.95,0.95,0.92}

\lstdefinestyle{mystyle}{
    backgroundcolor=\color{backcolour},   
    commentstyle=\color{codegreen},
    keywordstyle=\color{magenta},
    numberstyle=\tiny\color{codegray},
    stringstyle=\color{codepurple},
    basicstyle=\ttfamily\footnotesize,
    breakatwhitespace=false,         
    breaklines=true,                 
    captionpos=b,                    
    keepspaces=true,                 
    numbers=left,                    
    numbersep=5pt,                  
    showspaces=false,                
    showstringspaces=false,
    showtabs=false,                  
    tabsize=2
}

\newcommand*{\rn}{\textcolor{blue}}

\begin{document}
\title{\vspace*{1cm} LSTM-Based Forecasting and Analysis of EV Charging Demand in a Dense Urban Campus}

\author{
\IEEEauthorblockN{Zak Ressler\IEEEauthorrefmark{1},
Marcus Grijalva\IEEEauthorrefmark{2},
Angelica Marie Ignacio\IEEEauthorrefmark{3}, 
Melanie Torres\IEEEauthorrefmark{4},\\
Abelardo Cuadra Rojas\IEEEauthorrefmark{4},
Rohollah Moghadam\IEEEauthorrefmark{1},and
Mohammad Rasoul narimani\IEEEauthorrefmark{4}
}
\IEEEauthorblockA{\IEEEauthorrefmark{1}Electrical and Electronic Engineering Department, California State University-Sacramento, Sacramento, CA, USA}
\IEEEauthorblockA{\IEEEauthorrefmark{2}Electrical and Electronic Engineering Department, University of California, Riverside, Riverside, CA, USA}
\IEEEauthorblockA{\IEEEauthorrefmark{3}Computer Science Department, California State University, Northridge, Northridge, CA, USA}
\IEEEauthorblockA{\IEEEauthorrefmark{4}Electrical and Computer Engineering Department, California State University, Northridge, Northridge, CA, USA \\
Email: zakressler@csus.edu, arcusgrijalva27@gmail.com, angelica-marie.ignacio.883@my.csun.edu, \\melanie.torres.539@my.csun.edu, abelardo.cuadra-rojas.128@my.csun.edu, moghadam@csus.edu,rasoul.narimani@csun.edu}
}

\maketitle
%\conf{\textit{  Proc. of International Conference on Systems, Man, and Cybernetics (SMC 2025),  \\ 
%5-8 October 2025, Vienna - Austria}}

\begin{abstract}

This paper presents a framework for processing EV charging load data in order to forecast future load predictions using a Recurrent Neural Network, specifically an LSTM. The framework processes a large set of raw data from multiple locations and transforms it with normalization and feature extraction to train the LSTM. The pre-processing stage corrects for missing or incomplete values by interpolating and normalizing the measurements. This information is then fed into a Long Short-Term Memory Model designed to capture the short-term fluctuations while also interpreting the long-term trends in the charging data. Experimental results demonstrate the model's ability to accurately predict charging demand across multiple time scales (daily, weekly, and monthly), providing valuable insights for infrastructure planning, energy management, and grid integration of EV charging facilities. The system's modular design allows for adaptation to different charging locations with varying usage patterns, making it applicable across diverse deployment scenarios.
\end{abstract}


\section{INTRODUCTION}

The transition to electric vehicles (EVs) is crucial for mitigating climate change by reducing greenhouse gas emissions and reliance on fossil fuels. However, as EV adoption increases \cite{IEA2021}, the installation of numerous EV charging stations (EVCS) poses challenges to electric grids, particularly in dense communities. The increased demand for EVCS strains electric grid systems, leading to issues such as voltage drops and transformer overloads. Understanding these problems and their impacts is crucial for optimizing grid performance and ensuring sustainable EV infrastructure development. Therefore, accurately predicting EVCS load demand helps manage grid load, improve power network efficiency, and ensure reliable customer access to charging stations. Load prediction methods come in several forms, the first being models which do not use a deep neural network.

Traditional load forecasting methods, such as support vector regression, often struggle to capture the complex temporal dependencies and non-linear patterns inherent in EV charging behavior \cite{sun2016}. These patterns are influenced by numerous factors including time of day, day of week, seasonal variations, and charging station location. AI-based techniques have been developed for load forecasting, but conventional methods like artificial neural networks (ANNs), autoregressive integrated moving average (ARIMA), and fuzzy logic systems typically fall short when handling the long-term dependencies crucial for accurate EV charging prediction \cite{zhou2020}. These traditional approaches tend to suffer from vanishing gradient problems when dealing with extended sequences, and they often require extensive feature engineering to achieve reasonable performance. In contrast, deep learning architectures, particularly recurrent neural networks like LSTM, offer inherent advantages through their memory cells and gating mechanisms that can retain information over long sequences, automatically learn relevant features from raw data, and model the intricate relationships between multiple input and output variables simultaneously. In order to understand these advantages further, we explain the second approach to load forecasting.

Deep learning approaches offer substantial advantages for time series forecasting over traditional methods. Neural networks, particularly recurrent architectures like LSTM and gated recurrent unit (GRU), excel at capturing complex non-linear relationships and long-term dependencies without requiring extensive feature engineering \cite{hew2021}. These models can process multiple input variables simultaneously while accounting for their inter-dependencies, making them ideal for multivariate forecasting problems like EV charging demand prediction. Deep learning models show remarkable adaptability to changing patterns in the data, effectively handling the non-stationarity common in energy consumption time series \cite{seh2020}. Furthermore, transformer-based architectures have recently shown promising results by leveraging attention mechanisms to focus on the most relevant historical patterns for making accurate forecasts, especially for long-horizon predictions where traditional methods typically fail \cite{wen2023}. LSTM Models have emerged as powerful tools for performing time series forecasting across a wide range of domains. The field of power systems has seen great success using LSTMs for load forecasting, such as the experiments done by \cite{Weicong2024} who demonstrated LSTM superiority for day-ahead load forecasting for residential households. In the EV charging domain, a LSTM-based framework for predicting charging station demand was accomplished \cite{wevj2023}. Research has been done to explore using deep learning in order to predict peak load demand at particular locations to reduce EV charging load congestion \cite{Dou2023}. This combination of automatic feature learning, temporal pattern recognition, and adaptability to changing conditions makes deep learning approaches particularly well-suited for the challenges of modern forecasting applications. These deep learning techniques are the bedrock of this research paper and provide the key elements for the proposed LSTM architecture used for forecasting.

In this paper we propose a method to expand upon the LSTM architecture by providing multiple output forecasting to better assess trends of EV charging behavior. This approach addresses the fundamental issues regarding single output predictions being unable to accurately predict trend behavior and only attempt to approximate the next time step. The proposed approach trains the LSTM model to learn on multiple outputs ahead and approximate the trend to more accurately be applied to EV charging forecasting. Since forecasting for EV charging is meant to account for when peak demand will occur, having a model which reliably predicts trend behavior is crucial to addressing poor model performance in the standard single output LSTM.

{To verify the accuracy and applicability of the proposed forecasting framework, we developed a detailed model of the California State University, Northridge (CSUN) campus distribution grid using the PowerWorld Simulator. This digital representation captures the electrical topology of the campus, including transformer parameters, feeder configurations, distributed generation, and load connections across multiple buildings and parking structures~\cite{aguilar2024investigatingimpactelectricvehicle, iranpour2025automatedframeworkassessingelectric}. By integrating both the measured and LSTM-predicted EV charging demand profiles into this grid environment, we conducted comprehensive power flow simulations to evaluate how the predicted load affects system performance relative to the actual measurements. This validation step extends beyond numerical error metrics by assessing the physical consistency of the predicted load within a real distribution network model.}

{Simulation results demonstrated that voltage deviations between the predicted and actual load conditions remained extremely small across all representative buses, with maximum differences below 0.04\% ($\approx 4 \times 10^{-4} pu$). The close alignment between these two profiles confirms that the proposed LSTM-based model accurately captures both aggregate load dynamics and localized effects on grid behavior. These findings reinforce the reliability of the forecasting framework and its suitability for real-world deployment in campus-scale or urban distribution systems, where maintaining voltage stability and predicting demand fluctuations are critical for sustainable EV infrastructure integration.}

\section{PROBLEM STATEMENT}
Rising EVCS demand impacts the electric grid, causing voltage drops and transformer overloads that reduce equipment lifespan and compromise grid efficiency. Accurate load prediction is essential for optimizing grid performance and supporting reliable, sustainable EV infrastructure. This paper addresses the challenge by developing a precise prediction model for EV charging demand at CSUN, a dense urban campus. The proposed prediction model addresses short term temporal changes in daily charging while also forecasting broad change over multiple weeks. 

\begin{figure}[H]
    \centering
    \centerline{\includegraphics[width=0.9\columnwidth]{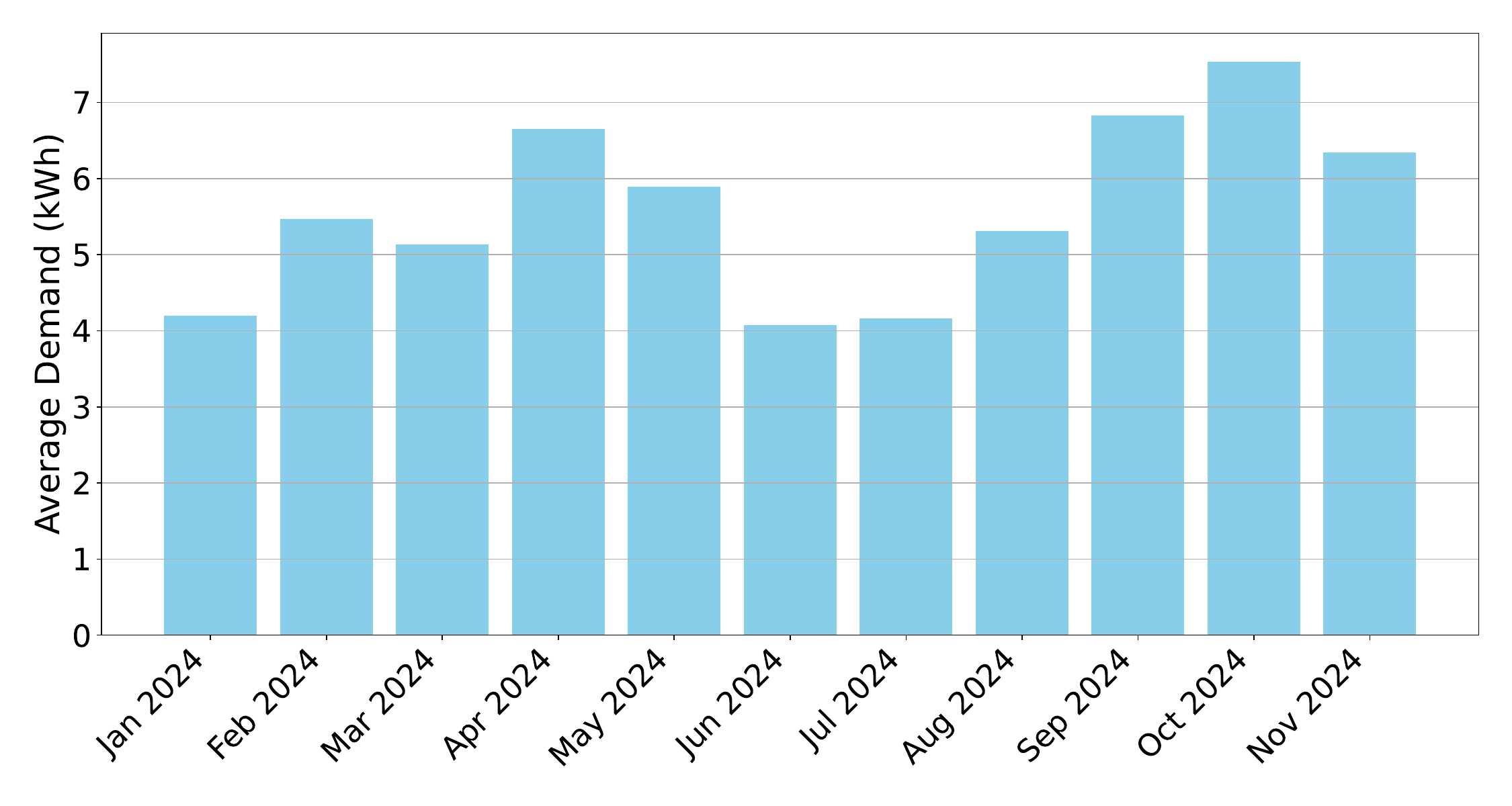}}
    \caption{Monthly Average Demand for Parking Garage B2 EV Charging Station}
    \label{fig:monthly-demand}
\end{figure}

\begin{figure}[H]
    \centering
    \centerline{\includegraphics[width=0.8\columnwidth]{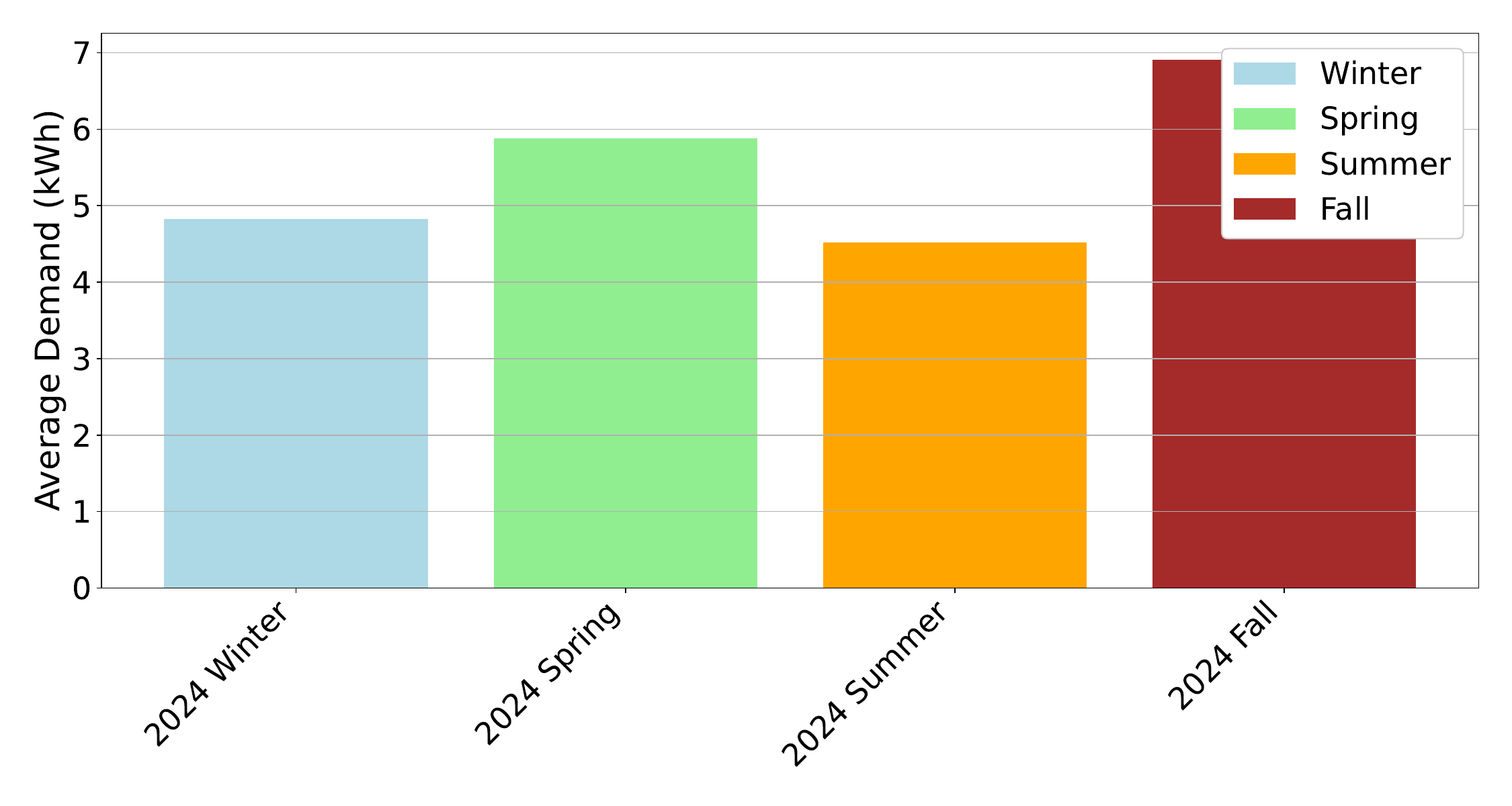}}
    \caption{Seasonal Average Demand for Parking Garage B2 EV Charging Station}
    \label{fig:seasonal-demand}
\end{figure}

This research provides a comprehensive methodological framework that transforms raw EV charging data into actionable forecasts through systematic pre-processing, feature extraction, and neural network modeling. The proposed approach leverages deep learning techniques to model these intricate relationships and produce accurate predictions across multiple time horizons. First, data is collected from multiple parking garage charging stations, followed by sophisticated data pre-processing and advanced LSTM modeling. In the data gathering phase, we collected high-frequency charging data from several public parking structures over a 12-month period, capturing daily and seasonal variations in charging patterns. The dataset consists of 15 minute charging intervals measuring the average, peak, and last measured kWh at a particular parking garage. Data for a single parking garage is selected based on specific criteria laid out in Section~\ref{sec:demandanalysis}. Once the data is collected, it is then sent through a pre-processing pipeline that handles missing values, normalizes features, and extracts relevant temporal patterns. Cross-correlation analysis and frequency domain transformations reveal underlying cyclical behaviors critical for accurate prediction. Through these signal processing techniques, our data analysis identified the dominant weekly patterns characterizing charging behavior. These processed inputs then feed into a carefully designed multi-layer LSTM architecture optimized for time series forecasting. Features such as the number of intervals the charger is used, the correlation between the intervals and peak usage, and the rate of change of the correlation are calculated and used as features for training the LSTM. The feature values are normalized and fed directly into the LSTM Model as input data. The data is divided into training, validation, and test sets for training and testing the LSTM Model. The LSTM Model undergoes gradient descent for training the model and uses the validation set with a set patience value for early stopping. Once the model has stopped training, the R-Squared, Mean Squared Error, Root Mean Squared Error, and Mean Absolute Error are calculated based on the test set.The architecture is optimized for capturing both short-term fluctuations and long-term dependencies in charging demand. The model leverages the engineered features extracted during the pre-processing phase, including daily averages, normalized counts, correlation signals, and their derivatives. By advancing the state of the art in EV charging load forecasting, this research contributes to the broader goal of sustainable transportation electrification while ensuring reliable and efficient energy system operation.

\section{EV CHARGING DEMAND ANALYSIS} \label{sec:demandanalysis}
This section presents an analysis of EV charging demand for CSUN with about 10 EV charging stations, distributed across the campus. According to available data, we chose one station with more data to perform analysis and prediction. 

\begin{figure}[H]
    \centering
    \centerline{\includegraphics[width=0.9\columnwidth]{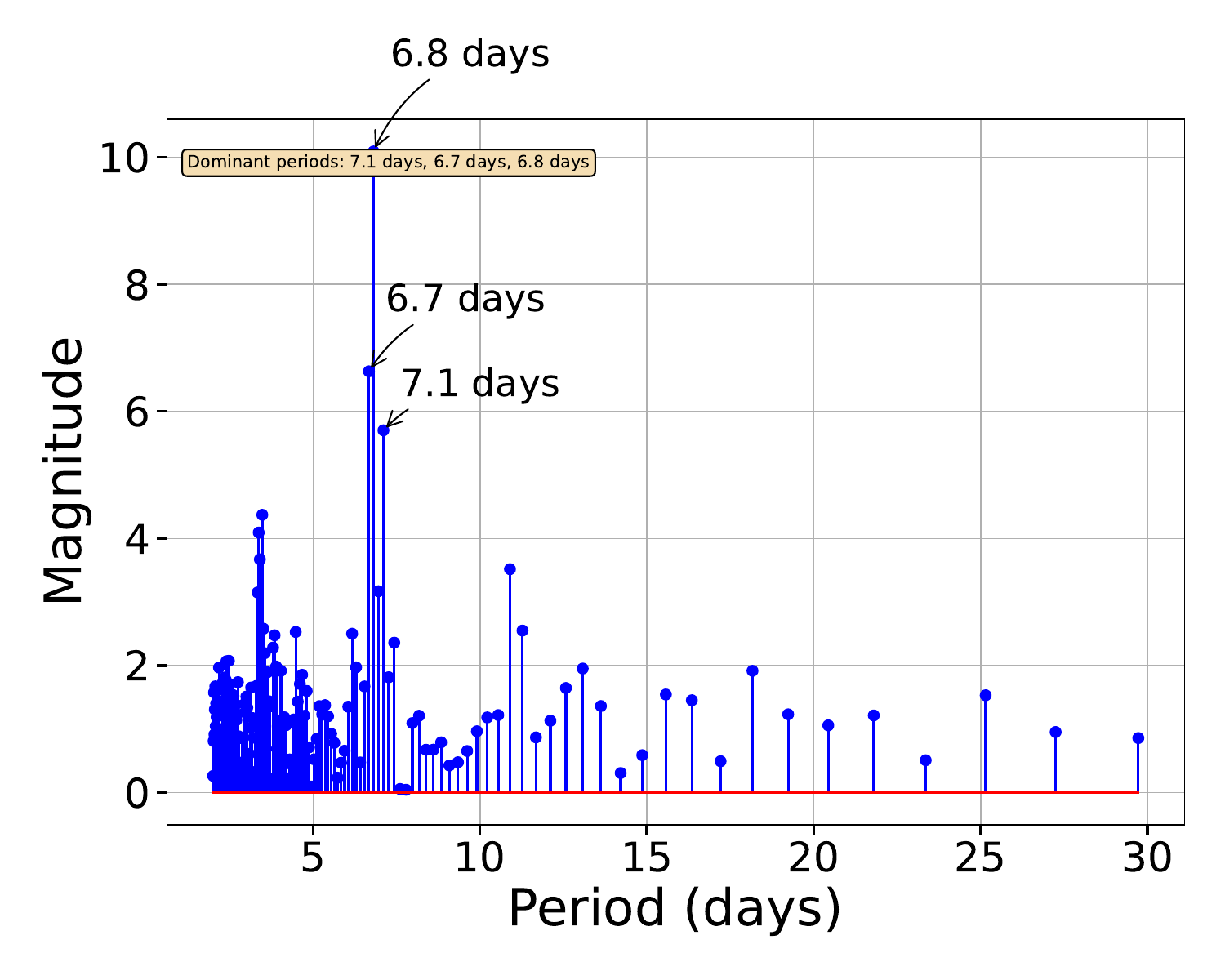}}
    \caption{Frequency Domain Analysis for Parking Garage B2 EV Charging Station}
    \label{fig:frequency-analysis}
\end{figure}

Figure \ref{fig:monthly-demand} displays the average daily demand for each month throughout the data collection period. The monthly resolution allows us to identify: \textbf{Month-to-Month Variations}: Fluctuations in charging demand between consecutive months, which may be related to academic schedules or seasonal activities on campus. \textbf{Annual Patterns}: Recurring patterns that appear in the same months across different years, such as reduced demand during summer breaks or increased demand during academic semesters. \textbf{Anomalous Months}: Unusual spikes or drops in demand that deviate from typical patterns, potentially indicating special events, infrastructure changes, or data collection issues. The monthly analysis provides a balance between granularity and trend visibility, making it useful for medium-term planning and resource allocation for EV charging infrastructure. Figure \ref{fig:seasonal-demand} aggregates the daily demand data by season (Winter, Spring, Summer, Fall) for each year. This visualization highlights: \textbf{Seasonal Patterns}: Differences in charging demand across the four seasons, revealing how weather conditions and seasonal activities affect EV usage. \textbf{Year-over-Year Seasonal Trends}: Changes in seasonal patterns from one year to the next, which may indicate evolving usage patterns or increasing EV adoption. \textbf{Academic Calendar Effects}: The influence of the university's academic calendar on charging demand, with potential differences between academic terms and breaks.

\begin{figure}[H]
    \centering
    \begin{subfigure}[b]{0.24\textwidth}
        \centering
        \includegraphics[width=\textwidth]{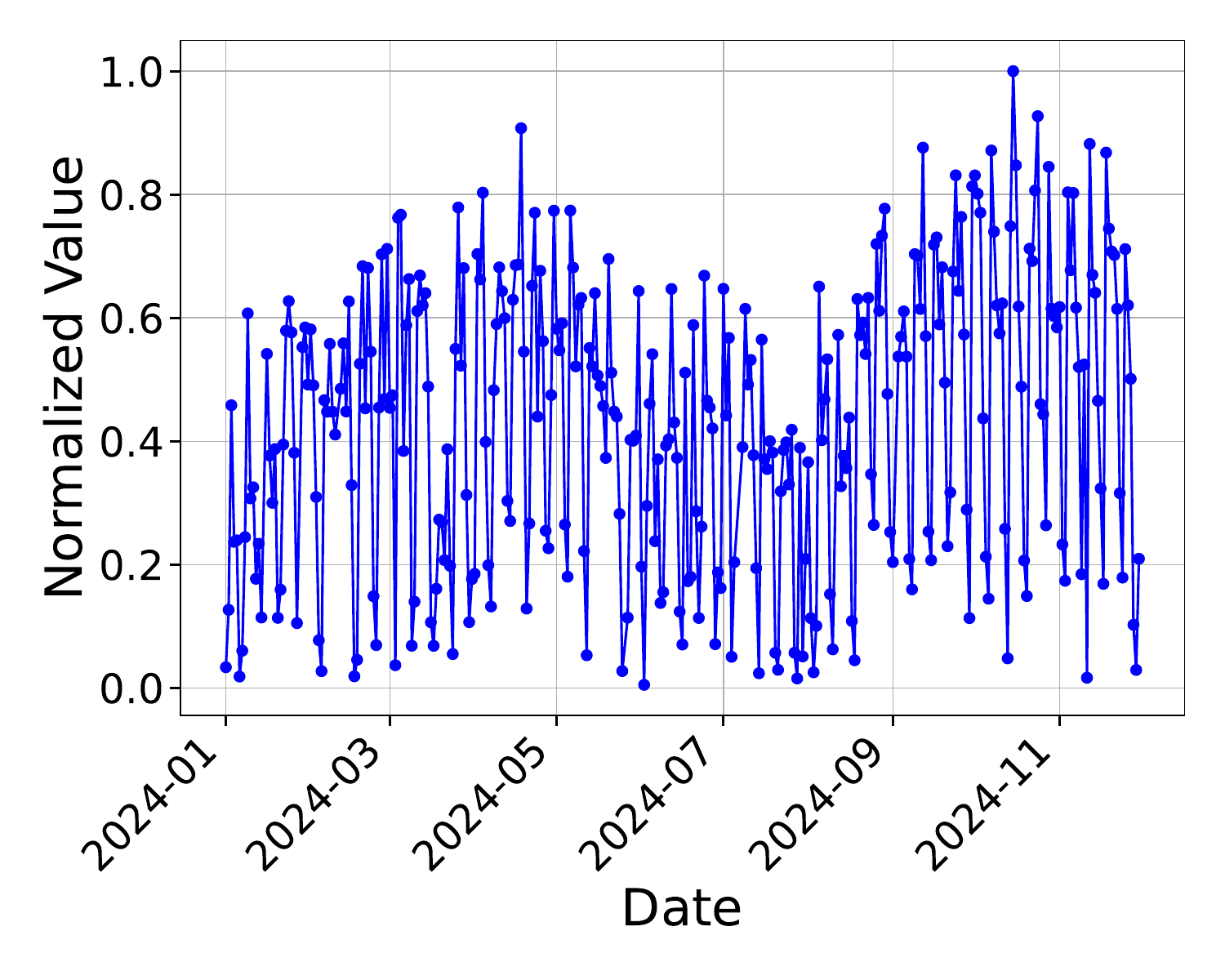}
        \caption{Normalized Daily Average}
        %fig_CrrAN_a}
    \end{subfigure}
    %\hspace{-0.5cm}
    \begin{subfigure}[b]{0.24\textwidth}
        \centering
        \includegraphics[width=\textwidth]{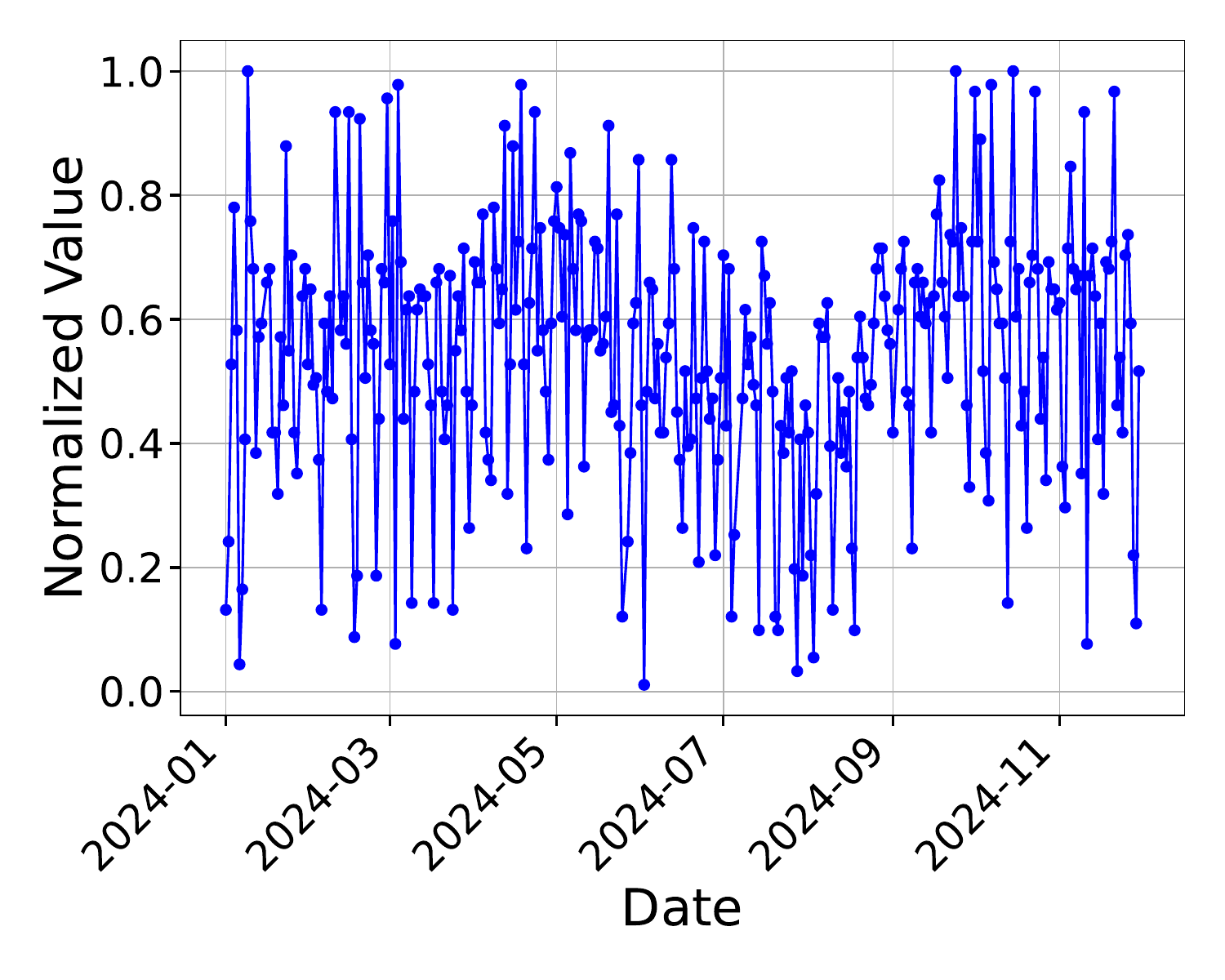}
        \caption{Normalized Non-zero Count}
        %fig_CrrAN_b}
    \end{subfigure}
    
    \vspace{0.1cm}
    
    \begin{subfigure}[b]{0.24\textwidth}
        \centering
        \includegraphics[width=\textwidth]{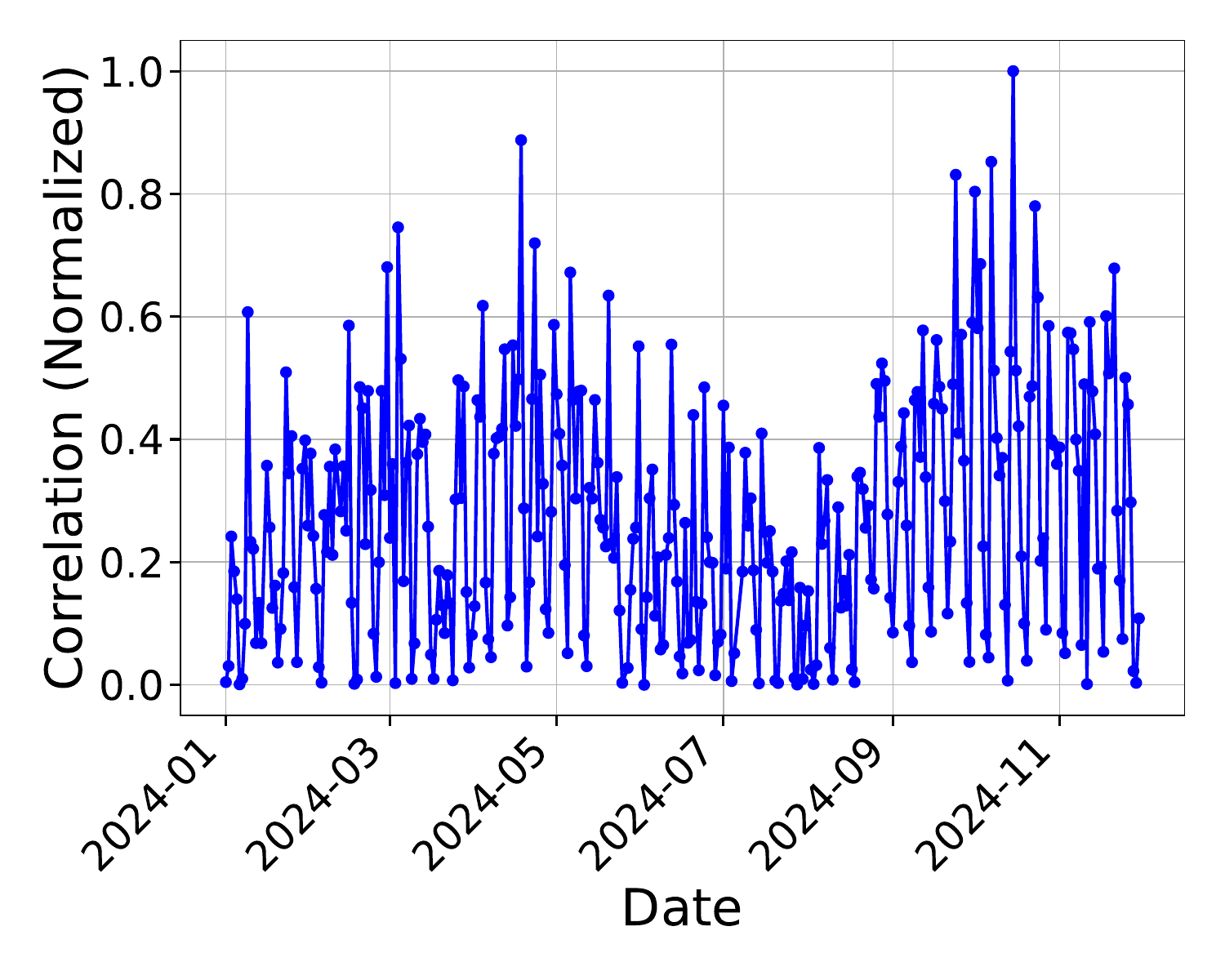}
        \caption{Cross-Correlation Signal}
        %fig_CrrAN_c}
    \end{subfigure}
    % \hspace{-0.5cm}
    \begin{subfigure}[b]{0.24\textwidth}
        \centering
        \includegraphics[width=\textwidth]{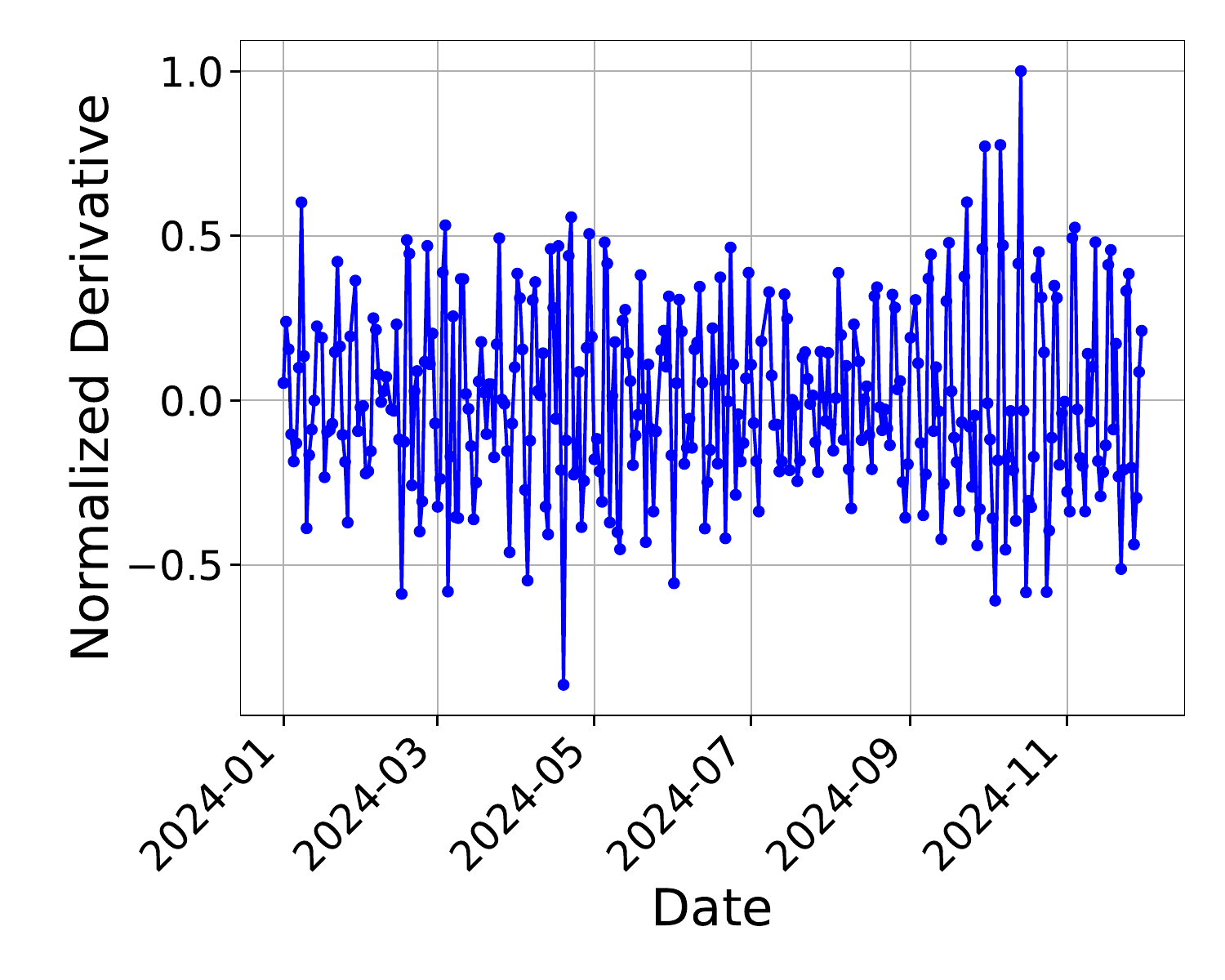}
        \caption{Derivative of Cross-Correlation Signal}
        %fig_CrrAN_d}
    \end{subfigure}
    \caption{Correlation analysis of EV charging patterns at Parking Garage B2: (a) Normalized daily average demand showing temporal patterns in charging intensity, (b) Normalized count of non-zero charging events reflecting frequency of station usage, (c) Cross-correlation signal revealing the relationship between charging frequency and demand intensity, and (d) Derivative of cross-correlation signal highlighting transition points in usage patterns.}
    \label{fig_correlation_analysis}
\end{figure}

% \begin{figure}
%     \centering
%     \centerline{\includegraphics[width=1\columnwidth]{Figure/ParkingGarageB2CorrelationAnalysis.pdf}}
%     \caption{Correlation Analysis for Parking Garage B2 EV Charging Station. The pre-processing step visualizes each data point extracted from the raw data in 4 categories. The first 2 boxes of data show the counts for charging averages in a day while the second box represents the normalized counts for charging intervals in a day. The third box shows the correlation line between the daily average and non-zero count. The 4th box shows the correlation derivative to visualize any differences or potential anomalies}
%     \label{fig:correlation-analysis}
% \end{figure}

Figure \ref{fig_correlation_analysis} presents a comprehensive correlation analysis for the Parking Garage B2 charging station. The analysis is divided into four subplots:

\begin{enumerate}
    \item \textbf{Normalized Daily Average}: This subplot shows the normalized daily average of EV charging demand. This enables us to identify relative changes in demand regardless of absolute magnitude.
    
    \item \textbf{Normalized Non-zero Count}: This subplot displays the normalized count of non-zero charging events per day. A non-zero count represents the number of time intervals within each day where charging activity was recorded. This metric helps understand the distribution of charging events throughout the day.
    
    \item \textbf{Cross-Correlation Signal}: This subplot illustrates the cross-correlation between the normalized daily average and the normalized non-zero count. Cross-correlation measures the similarity between these two signals as a function of the displacement of one relative to the other. Strong correlation suggests that high daily averages tend to occur on days with many charging events.
    
    \item \textbf{Derivative of Cross-Correlation Signal}: This subplot shows the rate of change (derivative) of the cross-correlation signal. Peaks in the derivative indicate rapid changes in the correlation pattern, which may correspond to significant shifts in usage patterns.
\end{enumerate}

The daily data points in the EV charging analysis are structured around multiple normalized metrics that capture different dimensions of charging behavior: Nonzero Count tracks station utilization frequency, Daily Average measures overall consumption intensity, Daily Maximum identifies peak demand periods, Cross-Correlation Signal quantifies the relationship between frequency and intensity, and the Derivative highlights transition points in usage patterns. This standardized format allows for meaningful comparisons across different time periods and charging locations while preserving the essential characteristics of the underlying data, with all metrics are normalized to the $[0,1]$ scale to enable pattern recognition independent of absolute magnitude.

The proposed multi-faceted visualization approach leverages these structured data points to reveal critical relationships between charging frequency and demand patterns through several analytical mechanisms. Temporal alignment analysis directly compares normalized count and average demand to identify whether station occupancy drives energy consumption; pattern recognition through cross-correlation quantifies the strength of relationship between frequency and intensity; transition detection via derivatives pinpoints behavioral shifts in charging patterns; cyclical behavior identification reveals recurring patterns at daily, weekly, and seasonal scales; and anomaly detection highlights unusual events that diverge from established patterns.

The 7-day, 14-day, and 30-day rolling averages provide complementary insights into EV charging patterns by analyzing the data at different temporal resolutions. The 7-day rolling average effectively filters out daily fluctuations while preserving weekly patterns, making it ideal for identifying regular weekly cycles driven by commuting habits and workday/weekend differences—patterns that strongly influence short-term operational decisions like staff scheduling and daily energy procurement. The 14-day (bi-weekly) rolling average strikes a balance between short-term variability and medium-term trends, capturing bi-weekly patterns that may correspond to pay periods or academic schedules while smoothing out anomalies that could distort weekly averages, providing valuable input for medium-term planning horizons such as maintenance scheduling and energy bidding strategies. The 30-day rolling average reveals monthly and seasonal trends by eliminating shorter-term fluctuations, highlighting gradual shifts in charging behavior related to seasonal weather patterns, academic terms, or holiday periods, which supports long-term strategic decisions including infrastructure expansion, capacity planning, and annual budget allocation. Together, these multi-scale rolling averages create a comprehensive analytical framework that enables both tactical responsiveness and strategic foresight in EV charging infrastructure management.

Figure \ref{fig:frequency-analysis} presents a frequency domain analysis of the cross-correlation signal from the Parking Garage B2 charging station. This analysis is conducted using the Fast Fourier Transform (FFT) to decompose the time-series data into its constituent frequency components. The x-axis represents the period in days, while the y-axis shows the magnitude of each frequency component. Prominent peaks in this plot represent dominant cyclical patterns in the data. The analysis is limited to periods of up to 30 days for better visibility of short to medium-term patterns. Key features of this visualization include\textbf{Dominant Periods}: The plot highlights the top three periodic components in the data, annotated with their corresponding period lengths in days. These dominant periods represent the most significant cyclical patterns in charging behavior.\textbf{Weekly Patterns}: Peaks around 7 days indicate weekly patterns, which likely correspond to differences in charging behavior between weekdays and weekends. \textbf{Bi-weekly and Monthly Patterns}: Peaks around 14 and 28-30 days may indicate bi-weekly or monthly patterns, potentially correlating with pay periods or academic schedules at the university. This frequency analysis helps identify regular patterns in EV charging demand that might not be immediately apparent in the time-domain representation, providing insights for capacity planning and demand forecasting. Frequency analysis is crucial as LSTM training can be dependent on cyclical data being present. Cyclical trends are far easier for the model to predict the future for and knowing the period of this data can significantly improve training times. Frequency analysis in LSTMs was used by Kulanuwat and Lattawit \cite{kulanuwat2021} to predict tidal patterns in hydrological time series data. Their findings show that LSTMs perform better with periodic data as opposed to non-cyclic data.
\begin{figure}[H]
    \centering
    \begin{subfigure}[b]{0.40\textwidth}
        \centering
        \includegraphics[width=\textwidth]{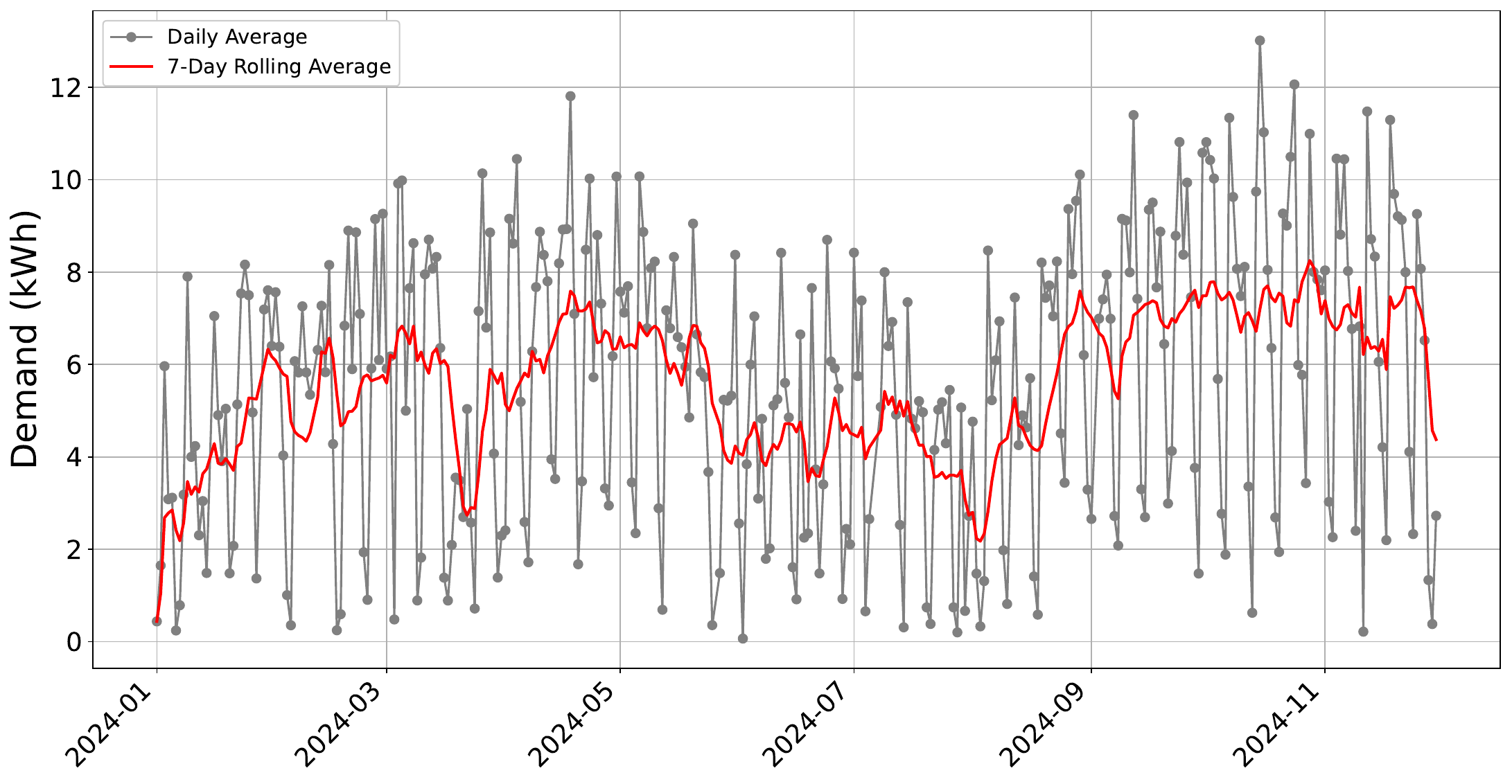}
        \caption{7 Day Rolling Average}
        %subfig-a}
    \end{subfigure}
    \begin{subfigure}[b]{0.40\textwidth}
        \centering
        \includegraphics[width=\textwidth]{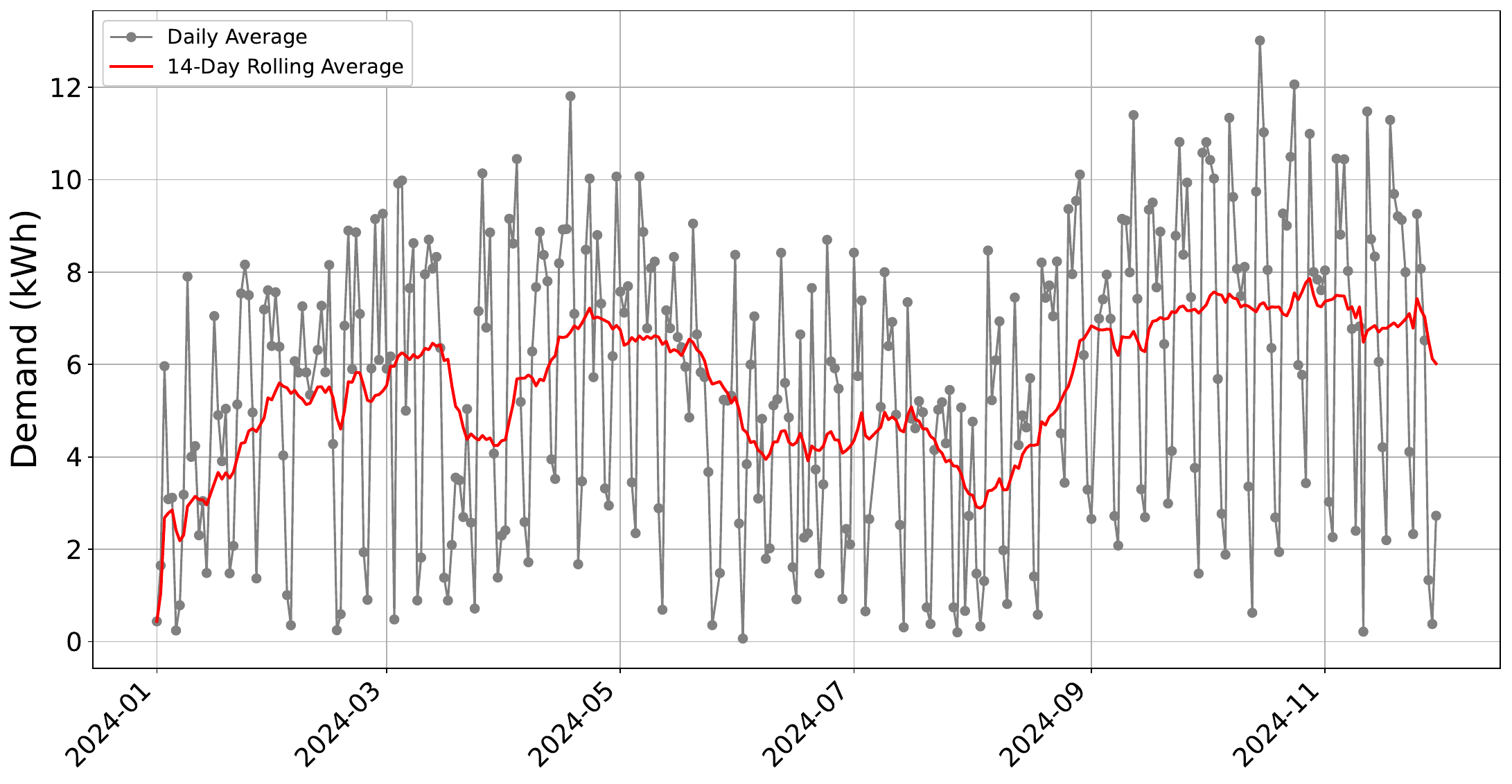}
        \caption{14 Day Rolling Average}
        %subfig-b}
    \end{subfigure}
    \begin{subfigure}[b]{0.40\textwidth}
        \centering
        \includegraphics[width=\textwidth]{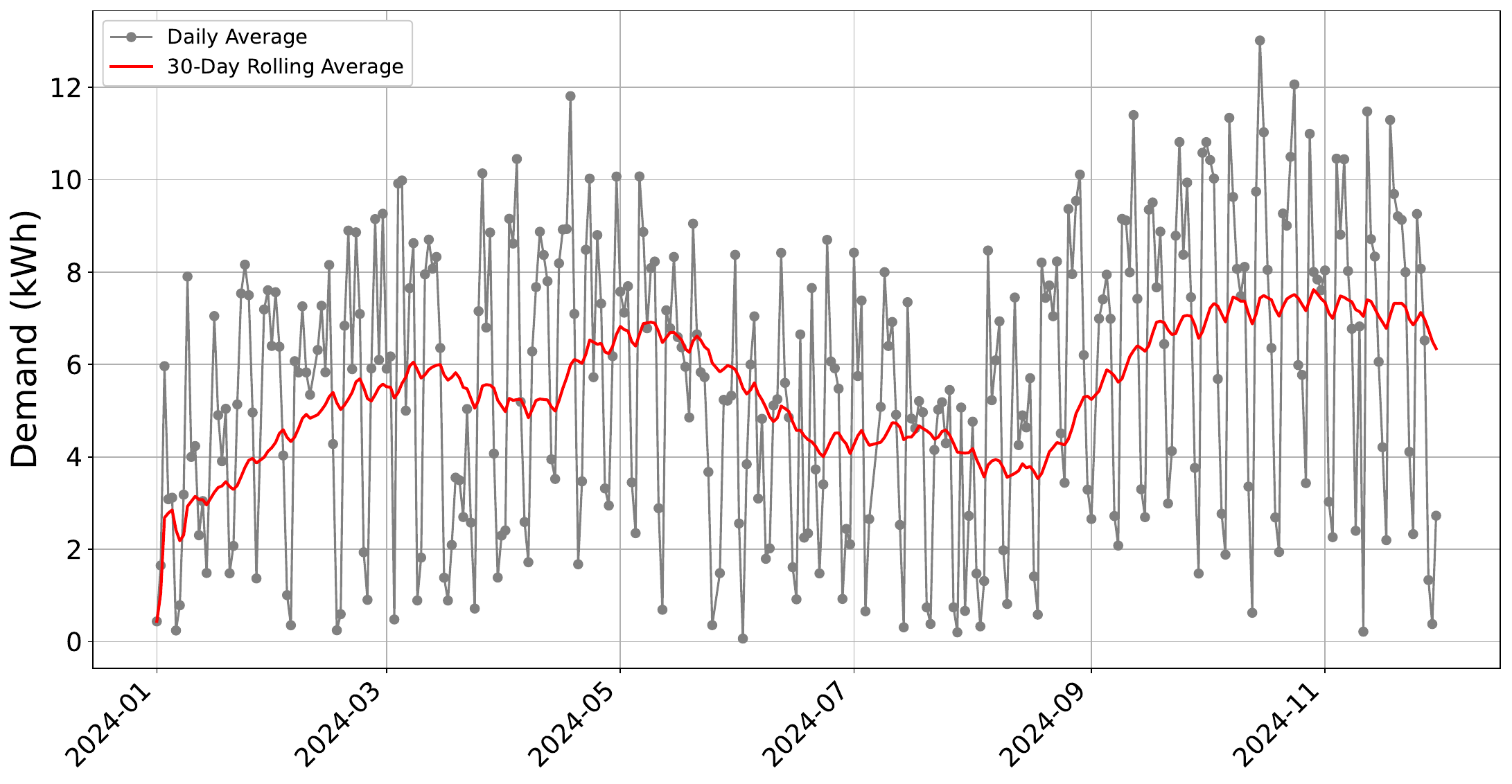}
        \caption{30 Day Rolling Average}
        %subfig-c}
    \end{subfigure}
    \caption{Rolling Average analysis of EV charging patterns at Parking Garage B2: (a) 7 Day Rolling Average, (b) 14 Day Rolling Average, (c) 30 Day Rolling Average}
    \label{fig:rolling-analysis}
\end{figure}
Figure \ref{fig:rolling-analysis} presents a time-series smoothing analysis using rolling averages with different window sizes for the Parking Garage B2 charging station. The figure consists of three subplots, each applying a different smoothing window to the daily average demand data:

\begin{enumerate}
    \item \textbf{7-Day Rolling Average}: The top subplot shows a weekly rolling average, which smooths out day-to-day fluctuations while preserving weekly patterns. This helps identify short-term trends in charging demand.
    
    \item \textbf{14-Day Rolling Average}: The middle subplot displays a bi-weekly rolling average, which further smooths the data to highlight medium-term trends while reducing the impact of weekly variations.
    
    \item \textbf{30-Day Rolling Average}: The bottom subplot presents a monthly rolling average, which reveals longer-term trends in charging demand by eliminating both daily and weekly fluctuations.
\end{enumerate}
In each subplot, the gray points and lines represent the raw daily average demand data, while the red line shows the smoothed trend using the respective rolling average window. This progressive smoothing approach helps distinguish between noise in the random day-to-day variations in charging demand. Regular patterns are more discernible at varying time scales making. The next section addresses the load prediction of EV charging demand.

\section{LONG SHORT TERM MEMORY (LSTM) FOR ELECTRIC VEHICLE (EV) CHARGING DEMAND PREDICTION}
The core functionality of the framework relies in the Long Short-Term Model which takes in a set of pre-processed data and outputs another set of predictions which follow the input data. The model is designed to predict multiple time steps ahead, with the output being reshaped to match the prediction horizon and feature dimensions. The advantages of using a neural network over a static algorithm for forecast prediction are as follows:
\begin{itemize}
    \item Adaptability: A Neural Network is able to reconcile new data in order to update its predictions based on new information not previously accounted for. 
    \item Complex Pattern Recognition: Neural Networks excel at classifying and predicting non-linear relationships in temporal data. LSTM Models are especially equipped for this task given their recurrent architecture.
    \item Scalability: A Neural Network with a larger dataset can extract more information about the data and make more accurate predictions. This allows a smaller model like the one presented in this paper to be expanded upon by providing more data and computation time.
\end{itemize}

\subsection{Overview of LSTM Networks}

\begin{figure}[H]
    \centering
    \centerline{\includegraphics[width=1\columnwidth]{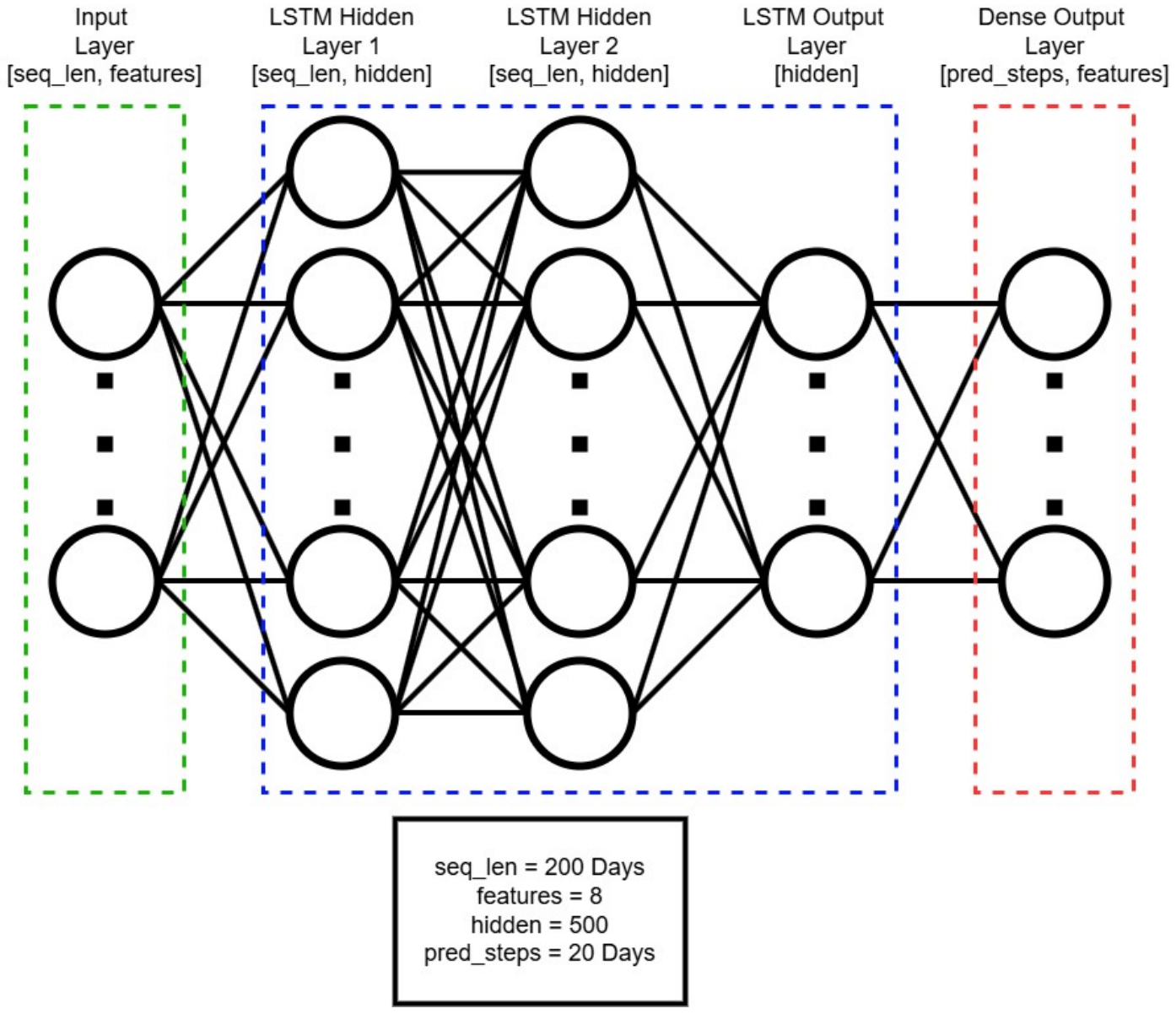}}
    \caption{The proposed customized LSTM architecture.}
    \label{fig:model-arch}
\end{figure}

% {\color{orange}{Brieftly talk about the LSTM Procedure - LSTM Block Diagram- Customized LSTM}}
% {\color{orange}{Add approach used in code vs how different from paper. Justify}}
The use of the LSTM and its advantages are laid out as initially proposed in \cite{Hoch1997}, who use the LSTM Model to solve the vanishing gradient problem with recurrent models. The proposed model uses the LSTM to avoid the vanishing gradient problem during training as well as utilize the recurrent architecture to analyze correlated time series trends in the EV charging data. The following equations describe this architecture as
\begin{eqnarray}\label{lstmEQ}
\begin{cases}
I_{\text{t}} = \sigma (X_{\text{t}}W_{\text{xi}}+H_{\text{t-1}}W_{\text{hi}}+b_{\text{i}}) \\
F_{\text{t}} = \sigma (X_{\text{t}}W_{\text{xf}}+H_{\text{t-1}}W_{\text{hf}}+b_{\text{f}}) \\
O_{\text{t}} = \sigma (X_{\text{t}}W_{\text{xo}}+H_{\text{t-1}}W_{\text{ho}}+b_{\text{o}})
% $\text{net}_{\text{out}_j}(t) = \sum_u w_{\text{out}_j u}y_u(t-1)$ \\
% $\text{net}_{c_j}(t) = \sum_u w_{c_j u}y_u(t-1)$ \\
% $s_{c_j}(t) = s_{c_j}(t-1) + y_{\text{in}_j}(t)g(\text{net}_{c_j}(t))$ for $t > 0$ \\
% $y_{c_j}(t) = y_{\text{out}_j}(t)h(s_{c_j}(t))$
\end{cases}\,,
\end{eqnarray}

where there are $h$ hidden units, the batch size is $n$
, and the number of inputs is $d$. The input is $X_{\text{t}} \in \mathbb{R}^{n\times d}$  and hidden state of the previous time step is $H_{\text{t-1}} \in \mathbb{R}^{n\times h}$. The input gate is $I_{\text{t}} \in \mathbb{R}^{n\times h}$, the forget gate is $F_{\text{t}} \in \mathbb{R}^{n\times h}$, and the output gate is $O_{\text{t}} \in \mathbb{R}^{n\times h}$.
% \begin{itemize}
$W_{\text{xi}}, W_{\text{xf}}, W_{\text{xo}} \in \mathbb{R}^{n\times h}$ and $W_{\text{hi}}, W_{\text{hf}}, W_{\text{ho}} \in \mathbb{R}^{h\times h}$ are  weighted parameters and $b_{\text{i}}, b_{\text{f}}, b_{\text{o}} \in \mathbb{R}^{1\times h}$ are bias parameters. The proposed LSTM model employs a sophisticated multi-input multi-output architecture that significantly extends beyond traditional single-step prediction approaches. The model processes sequences of multiple features simultaneously, enabling it to capture complex inter-dependencies across different dimensions of EV charging behavior. Rather than generating only the next time step prediction, the implementation forecasts multiple future time steps in parallel. This design allows the model to learn temporal patterns at varying timescales and maintain the contextual relationships between different features throughout the prediction horizon. Figure~\ref{fig:model-arch} in this paper demonstrates the proposed LSTM model's multi-input/multi-output capabilities by visualizing how the network processes all EV charging features simultaneously across the prediction horizon. The diagram shows the flow of input and hidden state information being concatenated and passed through each gate before arriving at the output gate. Unlike a single input and output model, the input data $X_{\text{t}}$ is batched to predict multiple time steps. The multi output model allows for better trend prediction than a standard single output model. Subsequent predictions in a multi-output model can be used to accurately form a trend line in the data rather than only predict a single point in the next time. 

% {\color{orange}REF NEEDED. ADD DIMENSIONS OF VECTORS / MATRICES.Formulas, dimensions, equations needs to be fixed (x)}

\subsection{Data Pre-processing for Load Prediction}
This section presents the process by which electrical load data is pre-processed from power values 
at individual locations into multiple sets of features useful for training the LSTM. This process is necessary for several reasons which affect the performance and training time of the model. The first major reason for pre-processing is for normalization and interpolation of missing values. Missing or abruptly changing values lead to worse performance in the LSTM. By setting upper and lower bounds in the data and normalizing those values, the model becomes easier to train and make more accurate predictions. Undoing the normalization process reverses this and allows us to make load predictions. The second major reason for pre-processing is feature extraction. The raw data contains more information about the behavior of the data that we can leverage to enhance the performance of the model. This involves calculating the gradient of the normalized data, the number of times the charging station recorded charging, the correlation between usage times and peak load times, etc. Each of these features gives the LSTM more data to train on when only given a small set of original raw data. This improves the models ability to generalize the data well and also indicates where the model may potentially fall short by misinterpreting the data.

For this study, we selected Parking Garage B2 as the primary focus of our analysis from among all available CSUN parking facilities. This selection was based on several critical factors that make B2 an ideal candidate for comprehensive EV charging pattern analysis. Parking Garage B2 exhibits exceptional data integrity with minimal missing values or anomalous readings compared to other campus locations. This completeness provides a robust foundation for reliable time-series analysis without requiring extensive interpolation techniques that might introduce bias. The B2 location demonstrates well-defined periodic charging patterns across multiple temporal scales. The charging patterns at B2 reflect typical academic usage scenarios while maintaining sufficient variation to test forecasting models effectively.

\subsubsection{Pre-processing Pipeline}
% The steps for how the data is pre-processed is divided into several steps:
% {\color{blue}{
% \begin{enumerate}
%     \item Data Loading and Cleaning (x)
%     \item Daily Usage Aggregation
%     \item Signal Processing
%     \item Frequency Analysis
%     \item Visualization
%     \item Output Generation
% \end{enumerate}}}

In this paper, the data pre-processing pipeline consists of a systematic six-stage approach designed to transform raw EV charging data into actionable insights. Initially, the Data Loading and Cleaning phase imports CSV files containing timestamped charging records, standardizes datetime formats, handles missing values through appropriate interpolation techniques, and filters anomalous readings that fall outside realistic power consumption parameters. Next, the Daily Usage Aggregation stage consolidates the high-frequency data into daily metrics, including nonzero charging event counts, daily average demand, and maximum power draw, providing a more manageable temporal resolution while preserving essential usage patterns. The Signal Processing phase then normalizes these metrics to enable meaningful comparisons and applies cross-correlation analysis to identify relationships between usage frequency and intensity, with gradient calculations to detect significant transitions in charging behavior. Following this, Frequency Analysis employs Fast Fourier Transform (FFT) techniques to decompose the time-series data into its constituent frequency components, revealing dominant cyclical patterns such as the weekly usage cycles. The Visualization stage generates multi-faceted graphical representations of the processed data, including correlation plots, FFT magnitude spectra, and rolling averages at various temporal scales. Finally, the Output Generation phase produces standardized data files containing the engineered features, normalized signals, and derived metrics that serve as inputs for the subsequent LSTM forecasting models.
% 1. 
% %\begin{itemize}
% %    \item Loads CSV files with EV charging data
% %    \item Converts timestamps to proper datetime format
% %    \item Handles missing values and zeros in the raw data
% %\end{itemize}

% 2. 
% %\begin{itemize}
% %    \item Groups data by day
% %    \item Calculates Daily average charging demand and Daily maximum charging demand
% %    \item Counts number of non-zero charging events per day
% %\end{itemize}

% 3. 
% %\begin{itemize}
% %    \item Normalizes data to facilitate comparison between different metrics
% %    \item Calculates cross-correlation between non-zero count and daily average
% %    \item Computes derivatives to analyze rate of change
% %\end{itemize}

% 4. 
% %\begin{itemize}
% %    \item Performs Fast Fourier Transform (FFT) on cross-correlation signals
% %   \item Identifies periodic patterns in the charging data
% %   \item Quantifies dominant cycles (weekly, monthly, etc.)
% %\end{itemize}

% 5. 
% %\begin{itemize}
% %    \item Raw data components (normalized non-zero count and daily average)
% %    \item Cross-correlation signals and their derivatives
% %   \item Frequency analysis showing periodic patterns
% %    \item Derivatives comparison to highlight trend changes
% %\end{itemize}

% 6. 
%\begin{itemize}
%    \item Saves processed data to the DataFrame for further analysis
%    \item Exports visualizations as image files
%\end{itemize}

This pre-processing pipeline transforms raw load data into actionable insights about usage patterns, periodic trends, and correlation between different metrics.

\subsubsection{Feature Extraction}

The feature extraction process consists of calculating several metrics from the pre-processed data for the purpose of training the LSTM. Values for the daily signal are extracted from the 15 minute intervals in the raw data using the following formulas
\begin{eqnarray}\label{sys_1}
\begin{cases}
NC(t) = \sum{i \in \text{day}_t} \mathbb{1}(Raw_i > 0)\\
DA(t) = \frac{1}{n_t} \sum{i \in \text{day}_t} Raw_i\\
DM(t) = \max{i \in \text{day}_t} Raw_i
\end{cases}\,,
\end{eqnarray}
where $NC(t)$, $DA(t)$, and $DM(t)$ are the Non-Zero Count, Daily Average, and Daily Maximum at time $t$, respectively. 

Values which are calculated are subsequently normalized for training using the following set of formulas

\begin{eqnarray}\label{sys_2}
\begin{cases}
NNC(t) = \frac{NC(t)}{\max(NC(t))}\\
NA(t) = \frac{DA(t)}{\max(DA(t))}\\
NM(t) = \frac{DM(t)}{\max(DM(t))}
\end{cases}\,,
\end{eqnarray}
where $NNC(t)$, $NA(t)$, and $NM(t)$ are the Normalized Non-Zero Count, Normalized Daily Average, and Normalized Daily Maximum at time $t$, respectively.

Correlation Metrics are quantified to show the relationship between the average signal and the non-zero count signal using the following formulas
\begin{eqnarray}\label{sys_3}
\begin{cases}
Crr(t) = NNC(t) \times NA(t)\\
CrrM(t) = NNC(t) \times NM(t)\\
CrrD(t) = \frac{dCrr(t)}{dt}\\
CrrMD(t) = \frac{dCorrM(t)}{dt}
\end{cases}\,,
\end{eqnarray}

where $Crr(t)$, $CrrM(t)$, $CrrD(t)$, and $CrrMD(t)$ are the correlation, the correlation maximum, the correlation derivative, and the correlation maximum derivative at time $t$ respectively.

Ratio and Ratio Maximum formulas are features extracted to measure the relationship between the count and average signals. This gives the LSTM model more ways of interpreting and predicting future data points. Below are the formulas
\begin{eqnarray}\label{sys_4}
\begin{cases}
R(t) = \frac{NNC(t)}{NA(t)}\\
RM(t) = \frac{NNC(t)}{NM(t)}
\end{cases}\,,
\end{eqnarray}
where $R(t)$ and $RM(t)$ are the ratio at time $t$ and ratio maximum at time $t$, respectively. The following frequency analysis formulas are used for determining periodicity in the parking garage correlation signal data. 

\begin{eqnarray}\label{sys_5}
\begin{cases}
X(f) = \left|\sum_{t=0}^{N-1} Crr(t) \cdot w(t) \cdot e^{-j2\pi ft/N}\right|\\
w(t) = \frac{1}{2} \left(1 - \cos\left(\frac{2\pi t}{N - 1}\right)\right)
\end{cases}\,,
\end{eqnarray}
Using these data processing formulas, we can outline the algorithm of the LSTM Model pipeline in Algorithm.

\begin{algorithm}
\caption{LSTM-based EV Charging Demand Forecasting}
\label{alg:lstm}
\SetAlgoLined
\KwData{Historical charging data $D = \{d_1, d_2, \ldots, d_T\}$, Sequence length $s$, Prediction horizon $h$, Noise level $\sigma$, Data multiplier $m$, Sampling Time $T$}
\KwResult{Predicted charging demand $\hat{D} = \{\hat{d}_{T+1}, \hat{d}_{T+2}, \ldots, \hat{d}_{T+h}\}$}

\textbf{Data preprocessing:}\;

Normalize data: $D_{\text{norm}} \leftarrow \text{NA}(D)$\;

Generate synthetic data: $D_{\text{synth}} \leftarrow \emptyset$\;

\For{$i = 1$ \KwTo $m$}{

    $D_i \leftarrow D_{\text{norm}} + \mathcal{N}(0, \sigma)$ \tcp*{Add Gaussian noise}
    
    $D_{\text{synth}} \leftarrow D_{\text{synth}} \cup D_i$\;
}

$D_{\text{train}} \leftarrow D_{\text{norm}} \cup D_{\text{synth}}$\;

\textbf{Create sequence samples:}\;

$X \leftarrow \emptyset, Y \leftarrow \emptyset$\;

\For{$t = s$ \KwTo $T-h$}{

    $x_t \leftarrow [d_{t-s+1}, d_{t-s+2}, \ldots, d_t]$\;
    
    $y_t \leftarrow [d_{t+1}, d_{t+2}, \ldots, d_{t+h}]$\;
    
    $X \leftarrow X \cup \{x_t\}, Y \leftarrow Y \cup \{y_t\}$\;
    
}

\textbf{Model training:}\;

Split data into training, validation, and test sets\;

Compile model with MSE loss and Adam optimizer ($lr = 0.00005$)\;

Initialize best validation loss $L_{\text{best}} \leftarrow \infty$\;

\For{epoch $= 1$ \KwTo max\_epochs}{

    Train model on $(X_{\text{train}}, Y_{\text{train}})$ with batch size $32$\;
    
    Compute validation loss $L_{\text{val}}$ on $(X_{\text{val}}, Y_{\text{val}})$\;
    
    \If{$L_{\text{val}} < L_{\text{best}}$}{
    
        $L_{\text{best}} \leftarrow L_{\text{val}}$\;
        
        Save model weights\;
        
    }
    \If{No improvement for $100$ epochs}{
    
        Early stop\;
        
    }
}

\textbf{Prediction generation:}\;

Load best model weights\;

$x_{\text{pred}} \leftarrow [d_{T-s+1}, d_{T-s+2}, \ldots, d_T]$ \tcp*{Last $s$ observations}

$\hat{D} \leftarrow \text{Model.predict}(x_{\text{pred}})$\;

Denormalize predictions: $\hat{D} \leftarrow \text{Denormalize}(\hat{D})$\;

\end{algorithm}

These formulas collectively quantify the temporal patterns, correlations, and cyclical behaviors in the EV charging data. The feature extraction process plays a crucial role in effectively modeling EV charging patterns. These carefully selected metrics capture different dimensions of charging behavior that are essential for accurate prediction.

The proposed metrics serve distinct analytical purposes that are vital for comprehensive EV charging demand forecasting. The count-based metrics (NC, NNC) quantify utilization frequency, revealing when charging stations experience high occupancy rates, which is critical for capacity planning and user access management. Intensity-based metrics (DA, NA, DM, NM) measure the power demand level, informing grid operators about potential peak loads and energy requirements. This dual perspective allows the model to distinguish between scenarios where many vehicles draw small amounts of power versus fewer vehicles drawing larger amounts—a critical distinction for grid stability management. In \cite{passalis2020} normalization was utilized to achieve better training performance on LSTM models for time series predictions. 

The correlation metrics (Crr, CrrM, CrrD, CrrMD) are particularly important as they capture the relationship between station utilization and power consumption intensity. These concepts are explored in \cite{bose2016} where a cross-correlation was used to classify neuromuscular diseases with deep learning models. The cross-correlation step is a crucial step in revealing underlying patterns in the data. These metrics reveal usage patterns that simple time series models would miss, such as when increased station occupancy doesn't proportionally increase power demand (indicating shorter charging sessions or lower-power vehicles). The derivative metrics add another dimension by quantifying the rate of change in these relationships, helping to identify transition points where charging behavior shifts—information that's invaluable for demand response program timing and dynamic pricing strategies. The ratio metrics (R, RM) provide normalized measurements of the relationship between count and intensity, allowing the model to learn patterns that persist across different absolute demand levels. Finally, the frequency analysis components enable identification of cyclical patterns (daily, weekly, seasonal) through Fourier transformation with Hanning window smoothing, which is essential for long-term forecasting and infrastructure planning. By incorporating this comprehensive set of metrics, the LSTM model can capture complex patterns across multiple time scales, leading to more accurate and nuanced predictions of EV charging demand.

It is important to note here that while frequency analysis is used to extract period information into a unique feature, the results are used in hyperparameter tuning rather than as input into the LSTM. The model architecture consists of 10 input features which capture the behavior of the raw charging data, including Nonzero Count, Daily Avg, Daily Max, Correlation, Correlation Derivative, Correlation Max, Correlation Max Derivative, Ratio, and Ratio Max. 

%The data processing pipeline performs several key transformations:
% {\color{orange}{ I think this part can be included in the algorithm I added to the paper (x) Added to Algorithm~\ref{alg:lstm}}}
% \begin{itemize}
%     \item Loading time-series data from CSV files with proper datetime parsing 
%     \item Normalizing features to ensure consistent scale across different metrics 
%     \item Creating sliding windows of appropriate sequence lengths for LSTM input 
%     \item Splitting data into training, validation, and testing sets with configurable ratios 
%     \item Batching data for efficient model training 
% \end{itemize}

% The architecture includes: 
% \begin{itemize} 
%     \item Multiple stacked LSTM layers for hierarchical feature extraction 
%     \item Linear output layer for prediction generation 
% \end{itemize}

% The model training incorporates: 
% \begin{itemize} 
%     \item Mean Squared Error (MSE) loss function for regression 
%     \item Adam optimizer with configurable learning rate 
%     \item Early stopping based on validation loss to prevent overfitting 
%     \item Dynamic batch sizing to accommodate different dataset characteristics 
% \end{itemize}

% Performance assessment includes: 
% \begin{itemize} 
%     \item Root Mean Square Error (RMSE) 
%     \item Mean Absolute Error (MAE) 
%     \item R-squared ($R^2$) coefficient 
%     \item 
% \end{itemize}
% }}

%{\color{red}{Visual comparison of predicted vs. actual charging patterns-Similar to what we have in Figure 3}- Add a table and include performance measures. We need to talk about them in simulation results. (x) }

The model is trained to learn the behavior of all 10 features per individual location. This allows the model to be tweaked per location for better results. Locations with too little data or chaotic data may affect the performance of the model and its ability to generalize to new data properly. By isolating each location, the performance of the model is much more accurate.

%\section{SOFTWARE DEVELOPMENT}
\subsection{LSTM Model Architecture}
This paper employs a multi-layer LSTM architecture implemented in PyTorch. The key components include:
\begin{itemize}
    \item Input layer accepting technical indicators
    \item 2 LSTM layers with configurable hidden dimensions
    \item Fully connected output layer for multi-step prediction
    \item Batch processing capabilities for efficient training
\end{itemize}

The model is designed to predict multiple time steps ahead, with the output being reshaped to match the prediction horizon and feature dimensions.

The core LSTM model implementation is shown below:

\subsubsection{Training Procedure}
The training process utilizes:
\begin{itemize}
    \item Mean Squared Error (MSE) loss function
    \item Adam optimizer with configurable learning rate
    \item Early stopping mechanism to prevent overfitting
    \item Training/validation/test split to evaluate model performance
    \item Batch training to manage memory usage efficiently
\end{itemize}

\subsubsection{Data Processing Pipeline}
We implement a specialized data generator class to create sequences suitable for training:

%\begin{lstlisting}[language=Python, caption=Data Processing Implementation]
%def create_sequences(self, data, seq_length, prediction_steps):
%        xs = []
 %       ys = []
 %       for i in range(len(data) - seq_length - prediction_steps + 1):
 %           x = data[i:(i + seq_length)]
 %           y = data[(i + seq_length):(i + seq_length + prediction_steps)]
 %           xs.append(x)
 %           ys.append(y)
 %       return np.array(xs), np.array(ys)
%\end{lstlisting}

\subsubsection{Prediction Generation}
For future predictions, the model:
\begin{enumerate}
    \item Takes the most recent \texttt{seq\_length} data points as input
    \item Generates predictions for the next \texttt{prediction\_steps} time points
    \item Visualizes the predictions alongside historical data
\end{enumerate}

The primary hyperparameters which significantly affect training include the sequence length, prediction steps, training size, validation size, added noise, noise magnitude, and patience. Sequence length and the prediction steps influence the patterns the model can infer. Smaller values train the model faster but reduce accuracy while larger values generally improve accuracy as seen in Table~\ref{tab:performance-metrics}.

\begin{figure}
    \centering
\includegraphics[scale=0.325,trim=0cm 0.0cm 0cm 0.0cm,clip]{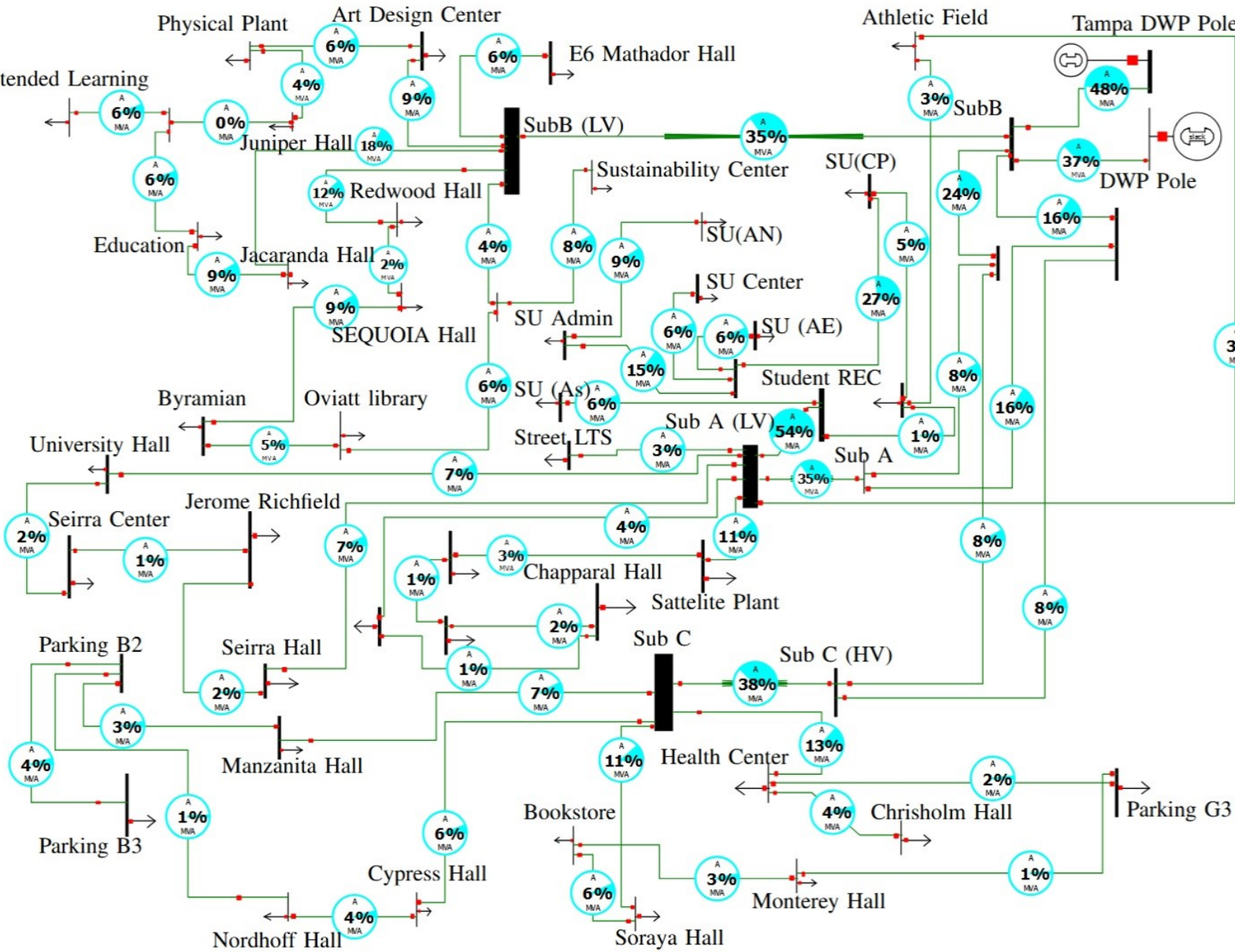}
    \caption{PowerWorld model of CSUN’s campus distribution network under regular (non-EV) load conditions. All feeders operate below thermal limits, indicating no line congestion in the baseline scenario.}
    \label{fig:csun_grid}
\end{figure}

\section{CSUN ELECTRIC GRID MODEL}
\begin{figure}
    \centering
\includegraphics[scale=0.23,trim=8.5cm 0.3cm 6cm 0.5cm,clip]{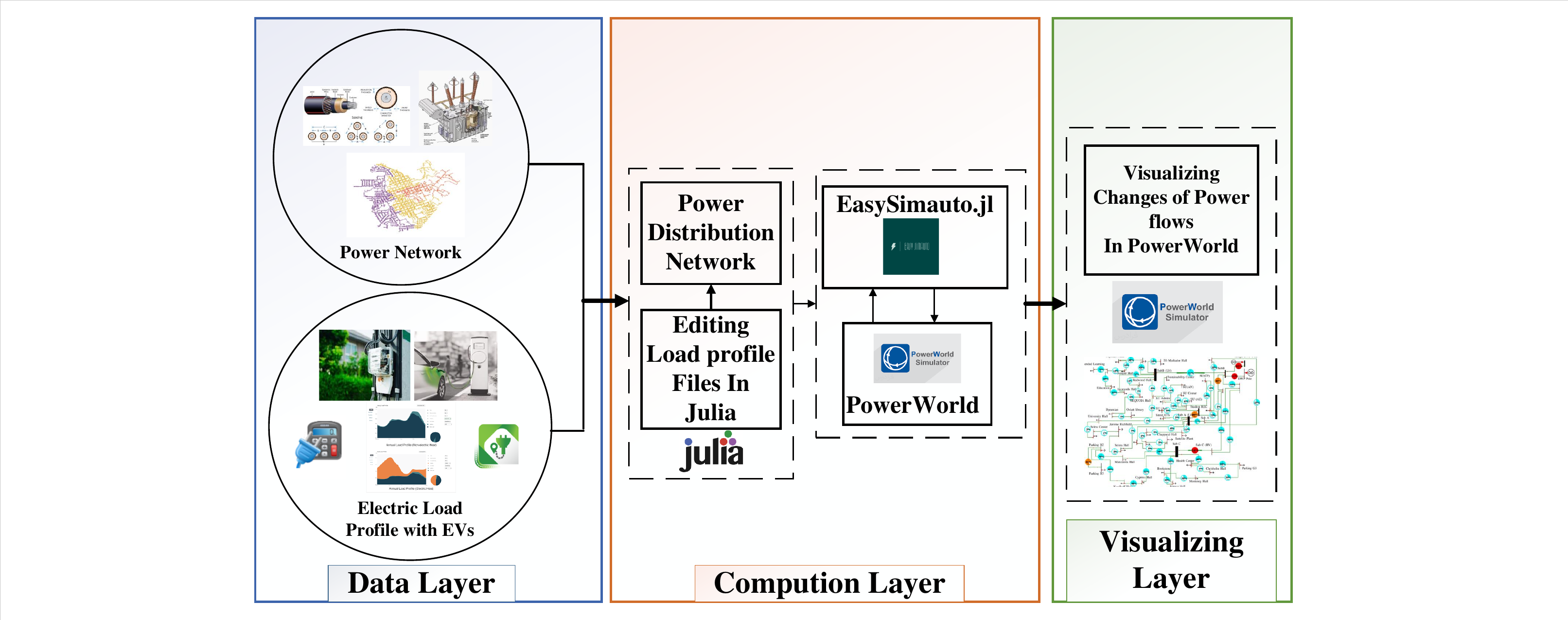}
    \caption{Schematic of the proposed framework for Dynamically Investigating the Impact of EV Chargers on Power Distribution Systems using Various tools and in three layers.}
    \label{fig:procedure}
\end{figure}

{To extend the forecasting analysis into system-level operation, we leveraged the detailed PowerWorld model of the CSUN distribution grid shown in Fig.~\ref{fig:csun_grid}. This model captures the physical topology and electrical parameters of the campus distribution system, including underground cables, transformers, distributed PV units, and representative feeders supplying academic, residential, and auxiliary loads. By faithfully incorporating system impedances, transformer nameplate data, and distributed generation profiles, the PowerWorld model serves as a digital twin of the CSUN campus grid, making it well-suited for time-series simulation of both real and predicted EV charging scenarios.
The simulation environment was further enhanced by integrating the automated analysis framework
 shown in Fig.~\ref{fig:procedure}. This framework consists of three interconnected layers: (i) the data layer, which organizes building load data and EV charging profiles (real or predicted); (ii) the computational layer, which uses SimAuto/Julia scripts to dynamically update PowerWorld simulations at each time step; and (iii) the visualization layer, which processes outputs such as voltage trajectories, line flows, and system losses into interpretable figures. This design enables consistent, repeatable analysis of different operating conditions and makes it possible to evaluate not only current loading patterns but also hypothetical growth scenarios or demand response strategies.}

{Using this setup, we conducted comparative studies by injecting both real measured EV charging data and LSTM-predicted demand profiles into the grid model. The primary goal was to determine whether forecast-based load profiles yield system responses comparable to those generated by actual data. By comparing bus voltages and line power flows, we assessed whether deviations between real and predicted datasets are significant enough to alter system-level decisions.}

\section{SIMULATION RESULTS}

The proposed LSTM-based forecasting system for EV charging demand demonstrates promising capabilities despite challenges inherent in the data. This section presents key findings and performance metrics from the implementation.

\subsection{Model Response to Data Characteristics}

The model demonstrated robust handling of several challenging data characteristics:

\subsubsection{Handling of Demand Spikes}

\begin{figure}[H]
    \centering
    \begin{subfigure}[b]{0.24\textwidth}
        \centering
        \includegraphics[width=1\columnwidth]{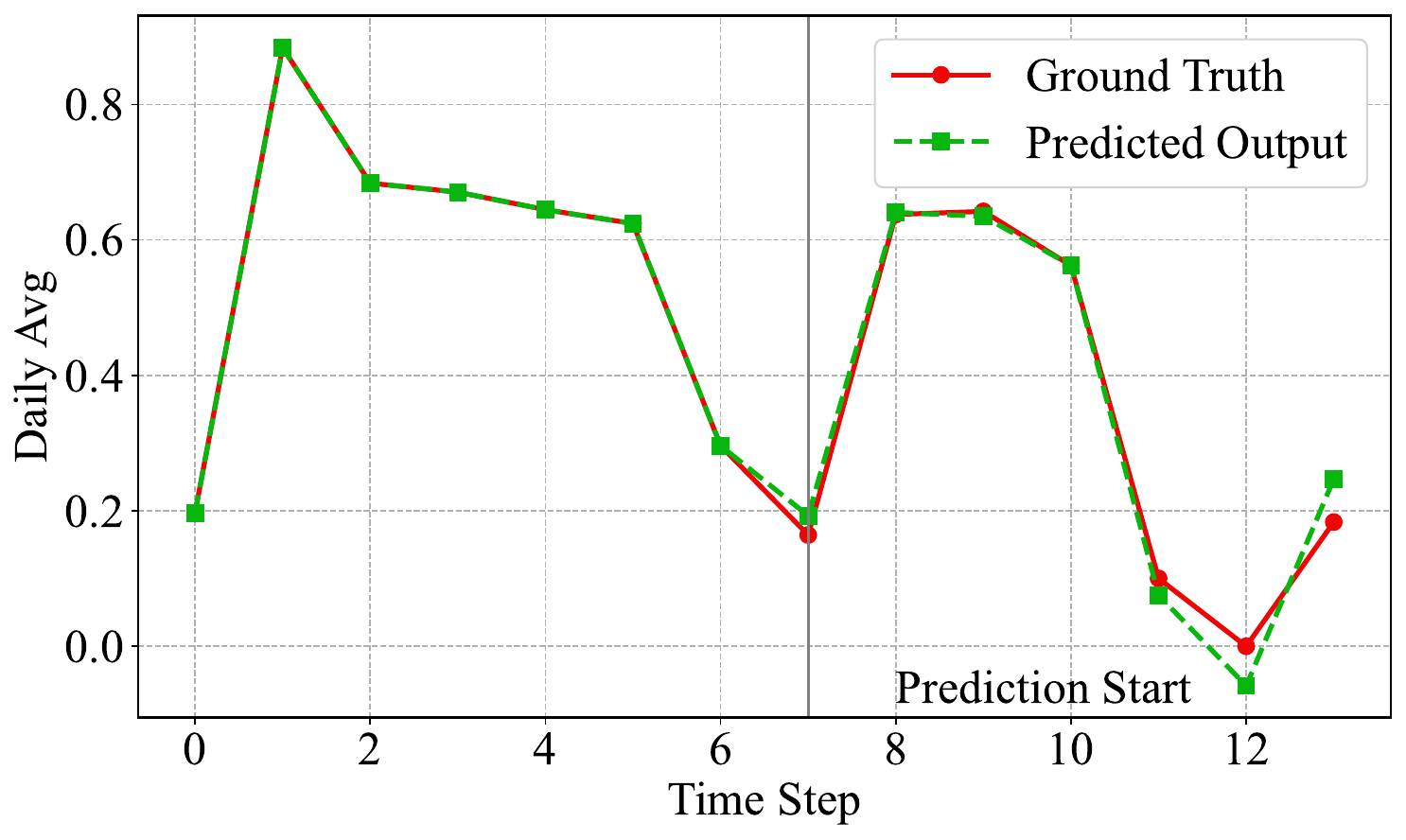}
        \caption{}
        %subfig-a}
    \end{subfigure}
    \begin{subfigure}[b]{0.24\textwidth}
        \centering
        \includegraphics[width=1\columnwidth]{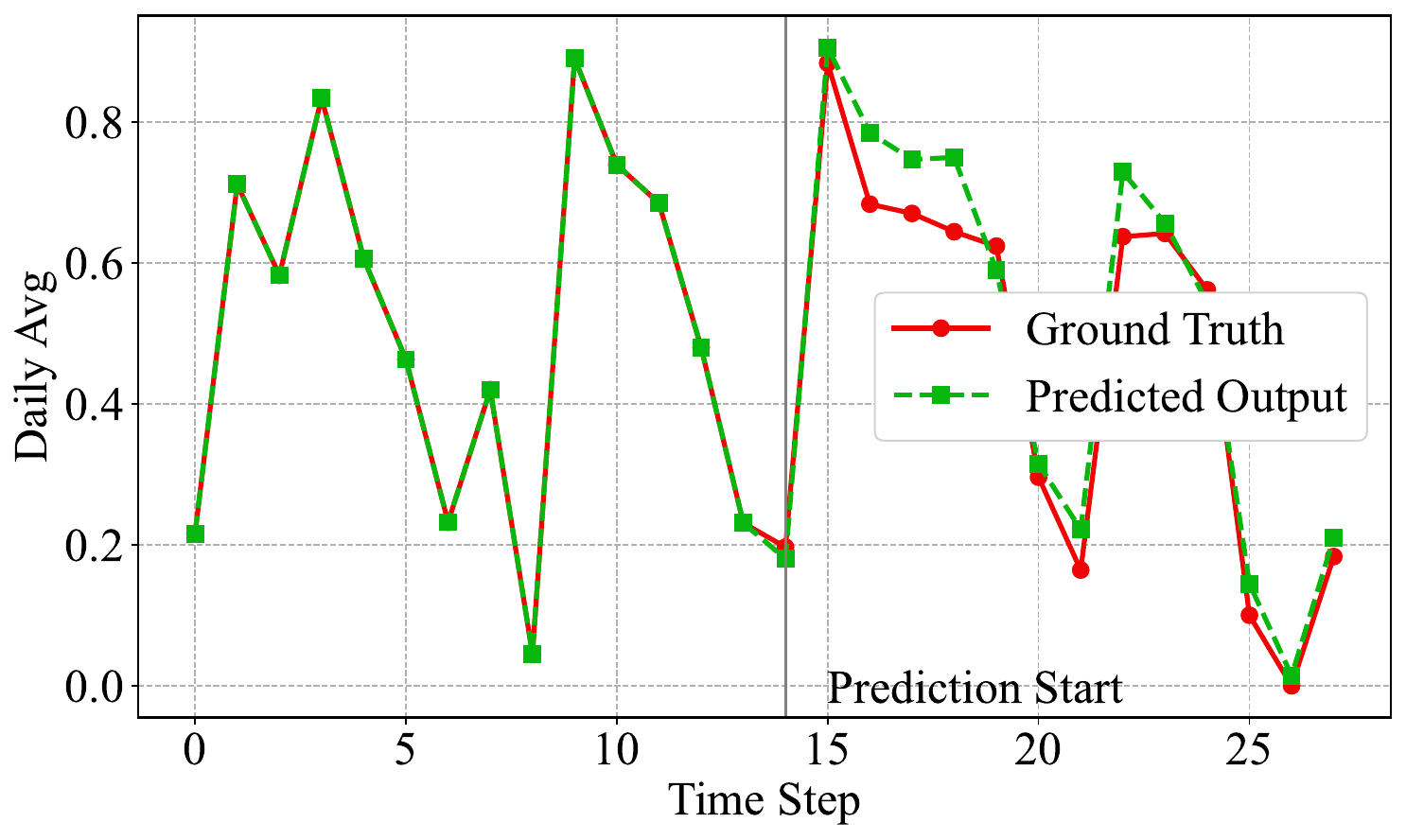}
        \caption{}
        %subfig-b}
    \end{subfigure}
    
    \vspace{0.1cm}
    
    \begin{subfigure}[b]{0.24\textwidth}
        \centering
        \includegraphics[width=1\columnwidth]{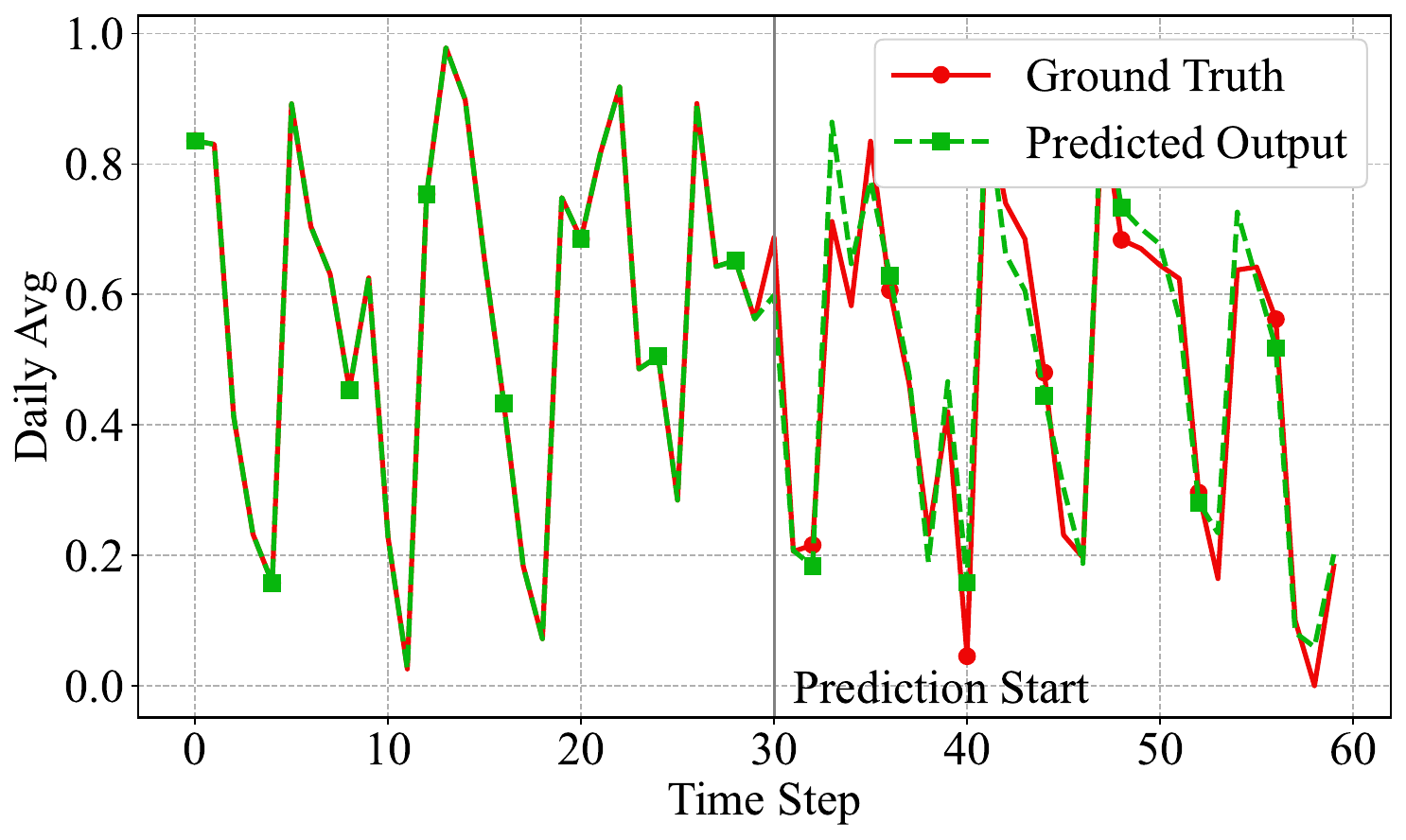}
        \caption{}
        %subfig-c}
    \end{subfigure}
    \begin{subfigure}[b]{0.24\textwidth}
        \centering
        \includegraphics[width=1\columnwidth]{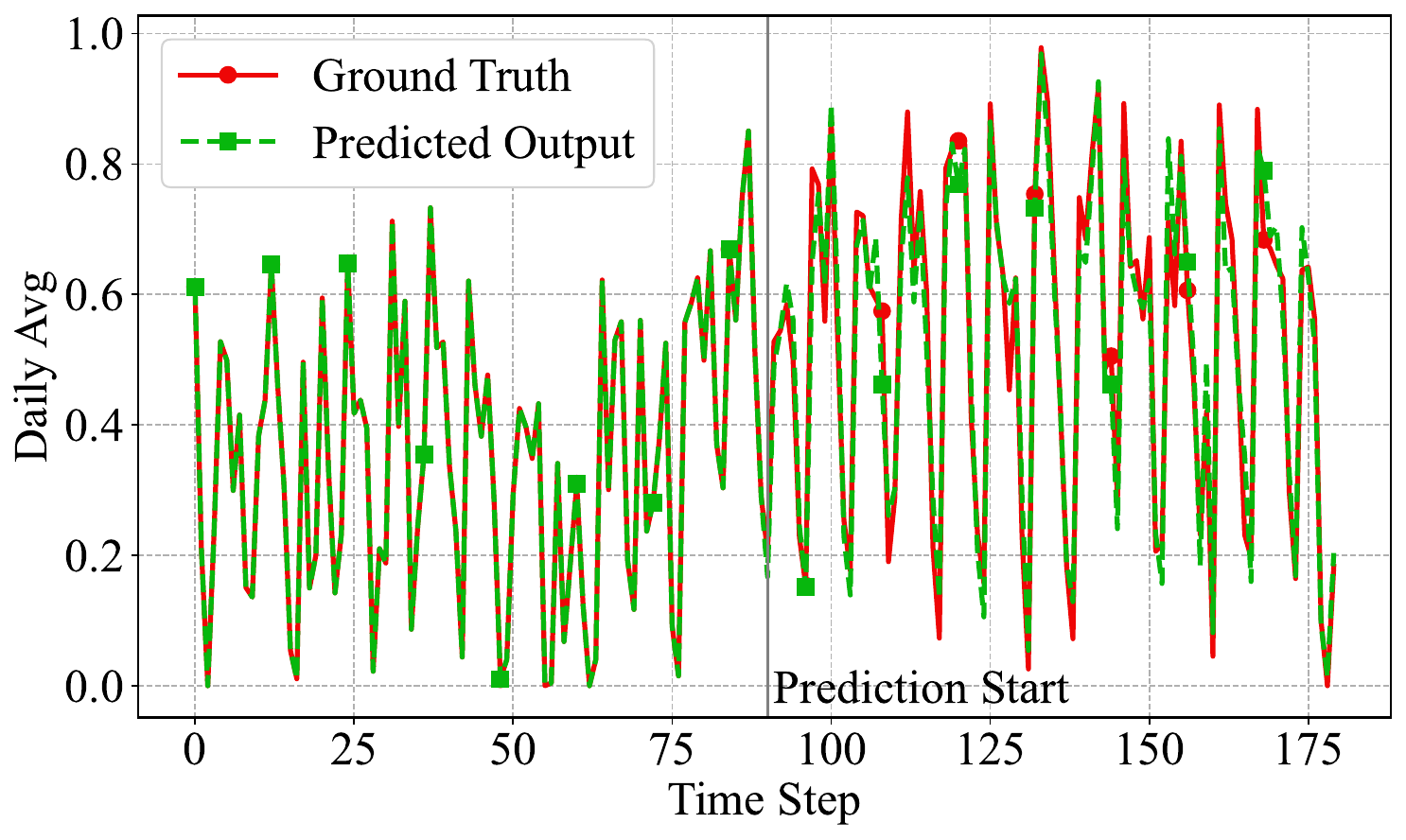}
        \caption{}
        %subfig-d}
    \end{subfigure}
    \caption{(a) 7 Day Daily Average Prediction vs Ground Truth (b) 14 Day Daily Average Prediction vs Ground Truth (c) 30 Day Daily Average Prediction vs Ground Truth (d) 90 Day Daily Average Prediction vs Ground Truth }
    \label{fig:daily-analysis}
\end{figure}

A key strength of the proposed implementation is its ability to handle demand spikes. Figure~\ref{fig:daily-analysis} illustrates how the model responds to sudden increases in charging demand over 30 days.

% \begin{figure}[ht] 
% \centering [Figure: Comparison of actual vs. predicted values during demand spike events] 
% \centerline{\includegraphics[width=1\columnwidth]{Figure/datapred.pdf}}
% \caption{Model prediction during charging demand spikes. The LSTM model anticipates demand increases with reasonable accuracy, though it sometimes underestimates the magnitude of extreme spikes.} \label{fig:spikes} \end{figure}

% {\color{orange}{\large COMMENTS 6-21-25:
% \begin{enumerate}
%     \item Prediction error is a lot here so as we discussed change the window and observe the behavior. This still needs improvement. (x)
%     \item Add noise to the current data and use it as training of 2023 year and then compare the results with the data we have from 2024 (x)
%     %\item Prediction is needed for seasonal, annual and monthly. Add a table and include the error metrics for each case. (x)
%     \item Figure 6 and Figure 7 still show major error. Add a graph showing error between the actual and predicted values. (x)
%     \item More simulations results are needed showing the effectiveness of the proposed approach.
%     \item Switch to blue and red color instead of green (x)
    
% \end{enumerate}
% }}

\begin{table}[htbp]
    \centering
    \caption{Optimal Hyperparameters for Different Multi Step Prediction Horizons}
    \label{tab:hyperparameters-multi}
    \begin{tabular}{lcccc}
        \toprule
        \textbf{Hyperparameter} & \textbf{Weekly} & \textbf{Bi-Weekly} & \textbf{Monthly} & \textbf{Seasonal} \\
        \midrule
        Sequence Length & 7 & 14 & 30 & 90 \\
        Prediction Steps & 7 & 14 & 30 & 90 \\
        Training Size & 0.6 & 0.6 & 0.6 & 0.6 \\
        Validation Size & 0.2 & 0.2 & 0.2 & 0.2 \\
        Test Size & 0.2 & 0.2 & 0.2 & 0.2 \\
        Learning Rate & 0.00005 & 0.00005 & 0.00005 & 0.00005 \\
        Batch Size & 32 & 32 & 32 & 32 \\
        Noise Level & 0.05 & 0.05 & 0.05 & 0.05 \\
        Multiplier & 10 & 10 & 10 & 10 \\
        Number of Epochs & 1000 & 10000 & 10000 & 10000 \\
        \bottomrule
    \end{tabular}
\end{table}

% \begin{table}[htbp]
%     \centering
%     \caption{Optimal Hyperparameters for Different Single Step Prediction Horizons}
%     \label{tab:hyperparameters-single}
%     \begin{tabular}{lcccc}
%         \toprule
%         \textbf{Hyperparameter} & \textbf{Weekly} & \textbf{Bi-Weekly} & \textbf{Monthly} & \textbf{Seasonal} \\
%         \midrule
%         Sequence Length & 7 & 14 & 90 & 270 \\
%         Prediction Steps & 1 & 1 & 1 & 1 \\
%         Training Size & 0.6 & 0.6 & 0.6 & 0.6 \\
%         Validation Size & 0.2 & 0.2 & 0.2 & 0.2 \\
%         Test Size & 0.2 & 0.2 & 0.2 & 0.2 \\
%         Learning Rate & 0.00005 & 0.00005 & 0.00005 & 0.00005 \\
%         Batch Size & 32 & 32 & 32 & 32 \\
%         Noise Level & 0.05 & 0.05 & 0.05 & 0.05 \\
%         Multiplier & 10 & 10 & 10 & 10 \\
%         Number of Epochs & 1000 & 10000 & 10000 & 10000 \\
%         \bottomrule
%     \end{tabular}
% \end{table}

\begin{table}[htbp]
    \centering
    \caption{Performance Metrics for Multi Step LSTM Forecasting Model}
    \label{tab:performance-metrics-multi}
    \begin{tabular}{lcccc}
        \toprule
        \textbf{Time Horizon} & \textbf{R²} & \textbf{MSE} & \textbf{RMSE} & \textbf{MAE} \\
        \midrule
        7 Days Ahead & 0.008 & 0.054 & 0.232 & 0.159 \\
        7 Day Ahead With Noise & 0.944 & 0.003 & 0.057 & 0.044 \\
        14 Days Ahead With Noise & 0.944 & 0.003 & 0.058 & 0.045 \\
        30 Days Ahead With Noise & 0.939 & 0.003 & 0.058 & 0.047 \\
        90 Days Ahead With Noise & 0.948 & 0.003 & 0.053 & 0.042 \\
        \bottomrule
    \end{tabular}
\end{table}

\begin{table}[htbp]
    \centering
    \caption{Performance Metrics for Single Step LSTM Forecasting Model}
    \label{tab:performance-metrics}
    \begin{tabular}{lcccc}
        \toprule
        \textbf{Time Horizon} & \textbf{R²} & \textbf{MSE} & \textbf{RMSE} & \textbf{MAE} \\
        \midrule
        7 Days 1 Day Ahead & 0.839 & 0.005 & 0.075 & 0.057 \\
        7 Day 1 Day Ahead With Noise & 0.876 & 0.006 & 0.074 & 0.055 \\
        14 Days 1 Day Ahead With Noise & 0.845 & 0.007 & 0.083 & 0.070 \\
        30 Days 1 Day Ahead With Noise & 0.907 & 0.004 & 0.064 & 0.048 \\
        90 Days 1 Day Ahead With Noise & 0.866 & 0.006 & 0.077 & 0.062 \\
        \bottomrule
    \end{tabular}
\end{table}

\begin{figure}[H]
    \centering
    \begin{subfigure}[b]{0.24\textwidth}
        \centering
        \includegraphics[width=1\columnwidth]{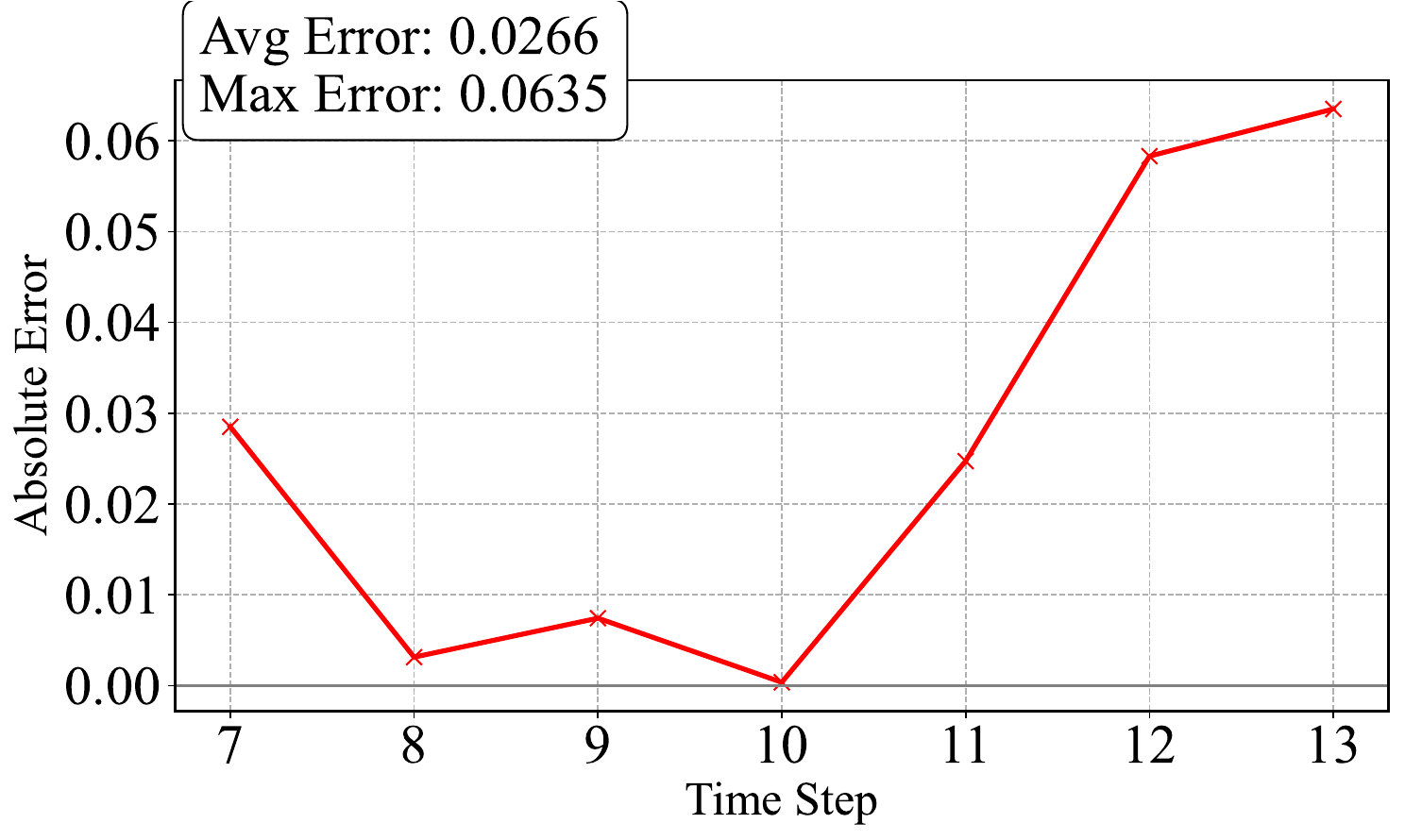}
        \caption{}
        %subfig-a}
    \end{subfigure}
    % \hspace{-0.5cm}
    \begin{subfigure}[b]{0.24\textwidth}
        \centering
        \includegraphics[width=1\columnwidth]{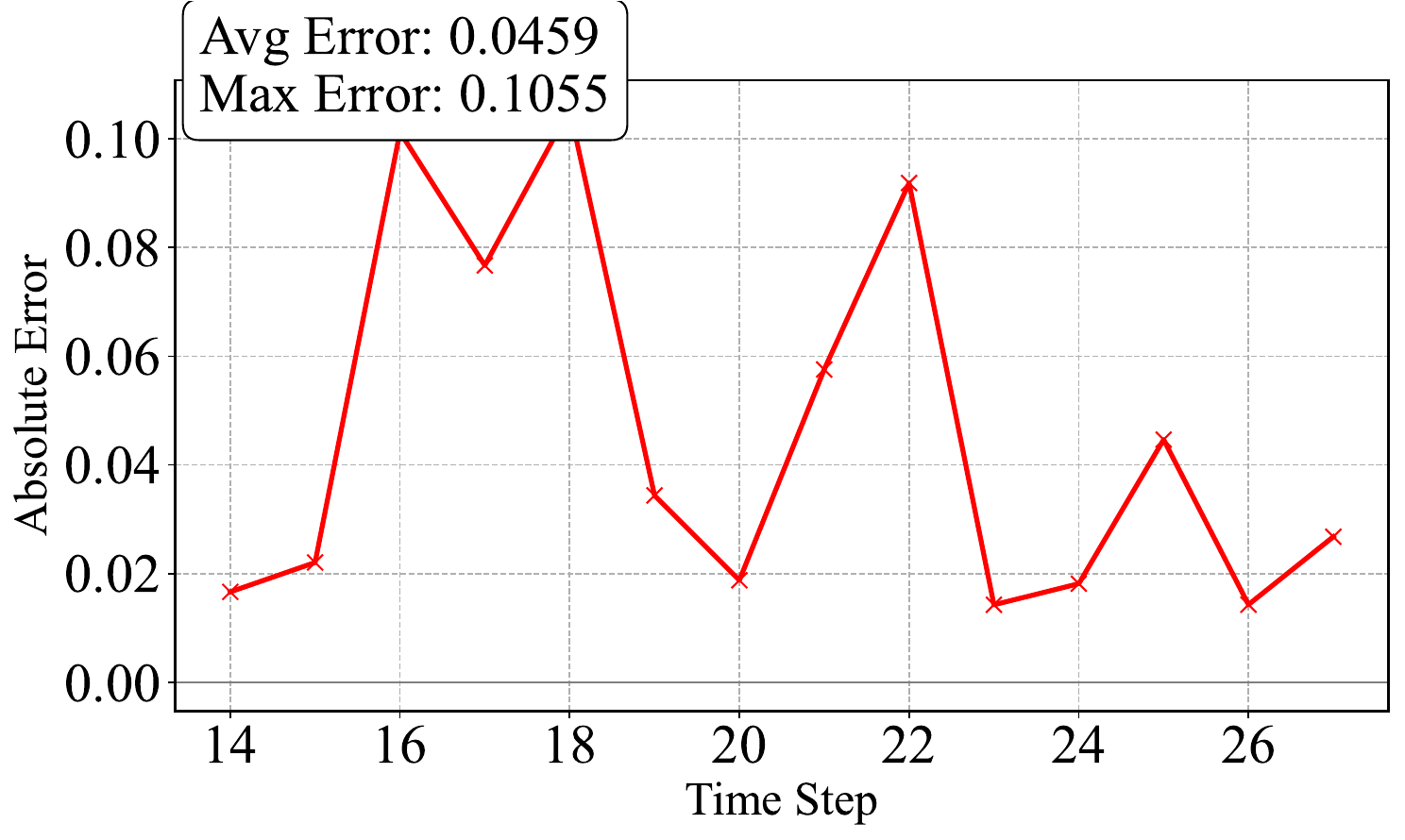}
        \caption{}
        %subfig-b}
    \end{subfigure}
    
    \vspace{0.1cm}
    
    \begin{subfigure}[b]{0.24\textwidth}
        \centering
        \includegraphics[width=1\columnwidth]{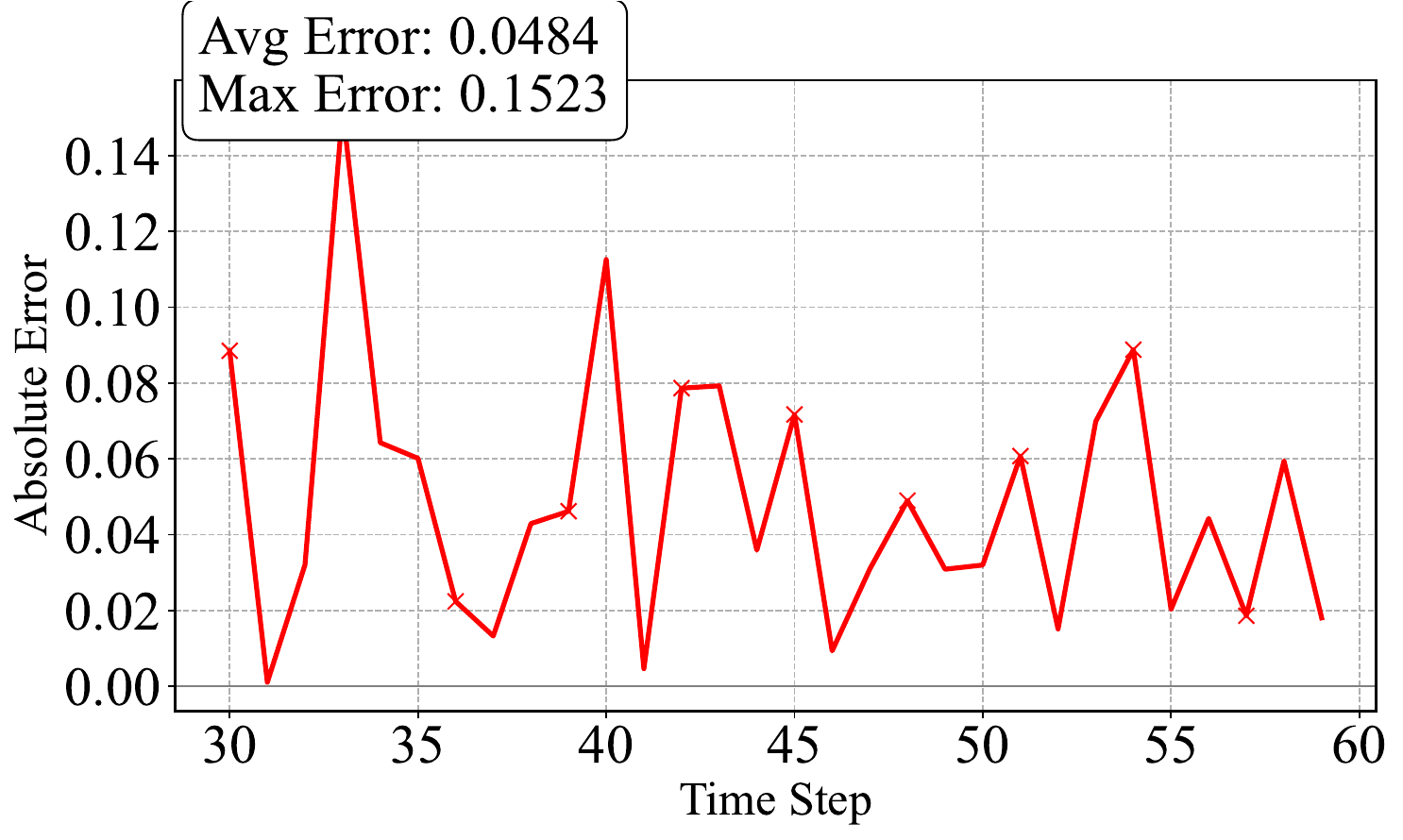}
        \caption{}
        %subfig-c}
    \end{subfigure}
    % \hspace{-0.5cm}
    \begin{subfigure}[b]{0.24\textwidth}
        \centering
        \includegraphics[width=1\columnwidth]{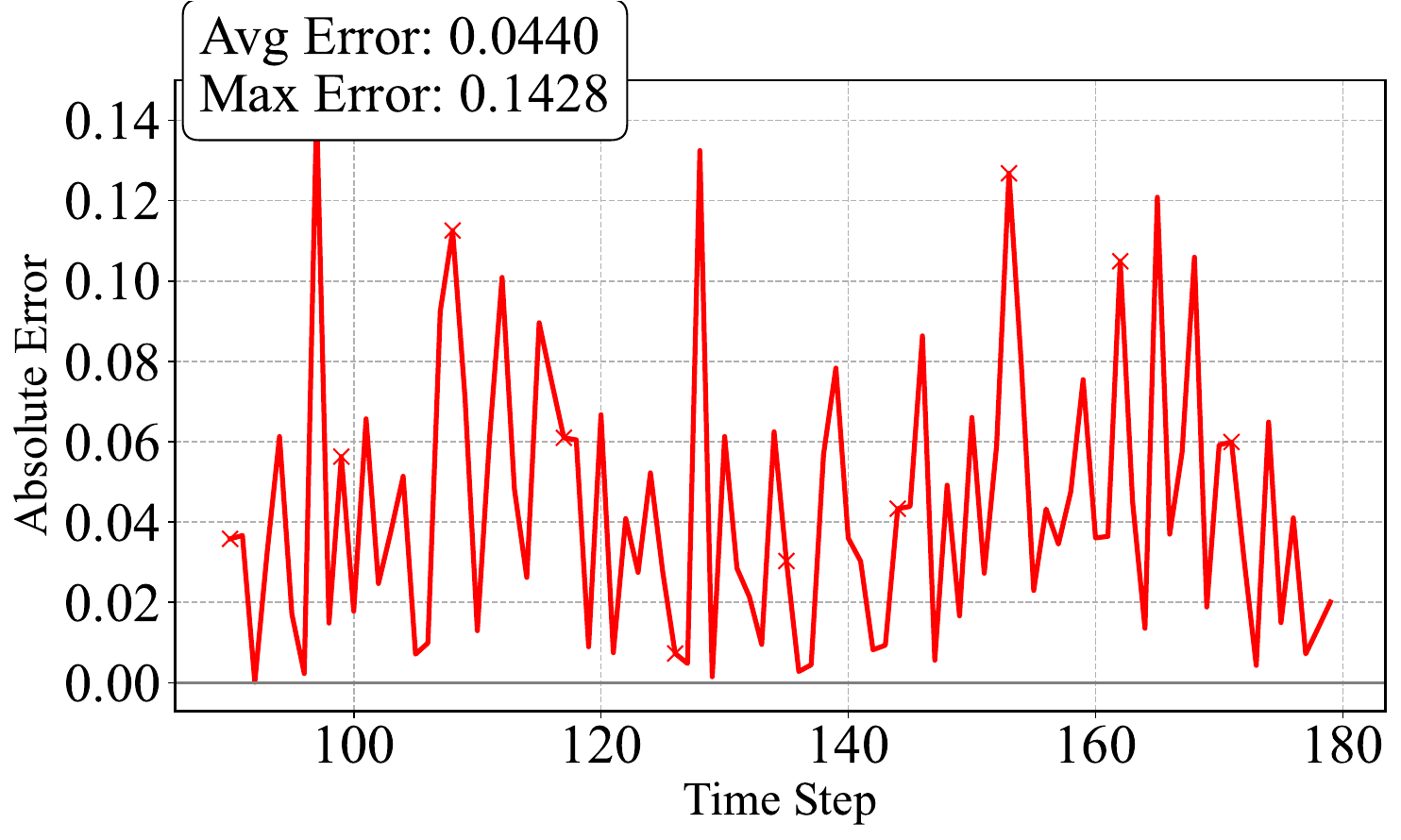}
        \caption{}
        %subfig-d}
    \end{subfigure}
    \caption{(a) 7 Day Absolute Error prediction (b) 14 Day Absolute Error prediction (c) 30 Day Absolute Error prediction (d) 90 Day Absolute Error prediction }
    \label{fig:daily-analysis-error}
\end{figure}

% \begin{figure}[ht] 
% \centering [Figure: 7 Day Ahead Prediction Error] 
% \centerline{\includegraphics[width=1\columnwidth]{Figure/B2_7_Day_Daily_Avg_error.pdf}}
% \caption{Model prediction error for short-term predictions given little previous data. The model error increases the further the predicted day is.} \label{fig:weekly-graph-error} \end{figure}

% \begin{figure}[ht] 
% \centering [Figure: 14 Day Ahead Prediction Error] 
% \centerline{\includegraphics[width=1\columnwidth]{Figure/B2_14_Day_Daily_Avg_error.pdf}}
% \caption{Model prediction error for biweekly cyclical demand. The model error appears to slightly decrease given further predictions.} \label{fig:biweekly-graph-error} \end{figure}

% \begin{figure}[ht] 
% \centering [Figure: 30 Day Ahead Prediction Error] 
% \centerline{\includegraphics[width=1\columnwidth]{Figure/B2_30_Day_Daily_Avg_error.pdf}}
% \caption{Model prediction error for monthly predictions given the previous month. The model error decreases the further the predicted day is.} \label{fig:monthly-graph-error} \end{figure}

% \begin{figure}[ht] 
% \centering [Figure: 90 Day Ahead Prediction Error] 
% \centerline{\includegraphics[width=1\columnwidth]{Figure/B2_90_Day_Daily_Avg_error.pdf}}
% \caption{Model prediction error for seasonal predictions given the previous season. The model error has a high variance but the average error remains lower than 14 and 30 day prediction models.} \label{fig:seasonal-graph-error} \end{figure}

While the model sometimes underestimates the magnitude of extreme spikes, it successfully captures the temporal pattern of these events, providing valuable advance warning of increased demand periods. An increase in accuracy comes from the addition of noise to the data and retraining with the new noise. The multiplier value with the noise inflates the dataset for better training. While increasing this value improves R2 accuracy, the model takes significantly longer to train and provides diminishing returns over time.

When comparing the performance metrics between the single and multi-step prediction, we can see a tangible improvement in the multi-step predictions. The R2 has between a 3\% and 10\% improvement in the R2 values from single to multi-step prediction across all metrics when noise is introduced into training. This is not the case for the prediction without noise, becoming significantly worse when going from single to multi-step prediction. This may be because further ahead predictions become more challenging for the model when working with less data. These findings are shown in Tables \ref{tab:performance-metrics-multi} and \ref{tab:performance-metrics}.
\subsection{Results on Electric Grid}
The LSTM-based forecasting model developed in this research has significant implications for electric grid management in the context of increasing EV adoption. As electric vehicles continue to proliferate, their charging demands introduce new challenges for grid stability, capacity planning, and energy distribution. The proposed predictive model addresses these challenges through accurate forecasting across multiple time horizons.
The ability to forecast EV charging demand with high accuracy (R2 = 0.939 for month-ahead predictions) enables grid operators to implement effective load balancing strategies. By anticipating demand spikes 24-72 hours in advance, operators can adjust generation schedules to accommodate predicted charging loads, implement dynamic pricing to incentivize off-peak charging, and activate demand response programs to shift non-critical loads during predicted high-demand periods.

\subsection{Discussion}

The follow need to be considered in the development procedure. When deploying the model on new EV charging data it is important to account for specific constraints on the dataset. A large sample of data is strictly necessary for accurate predictions on load and forecasts can range anywhere from 1 week to 3 months ahead depending on application. A limited set of data will inhibit the ability of the model to generalize properly. Table~\ref{tab:performance-metrics} shows the stark contrast in accuracy when 10 years of synthetic data with noise is added to the dataset as opposed to the dataset with only a single year of data points.
When deploying with a large dataset, another major consideration is proper hyperparameter tuning. Adjusting parameters like batch size and training size can significantly increase or decrease training times depending on the size and noise present in your data. It is recommended to use an optimizer like PSO to find ideal hyperparameter values for your model with your data. 

\begin{figure}
    \centering
\begin{tikzpicture}
\node at (0,0) {\includegraphics[scale=0.5]{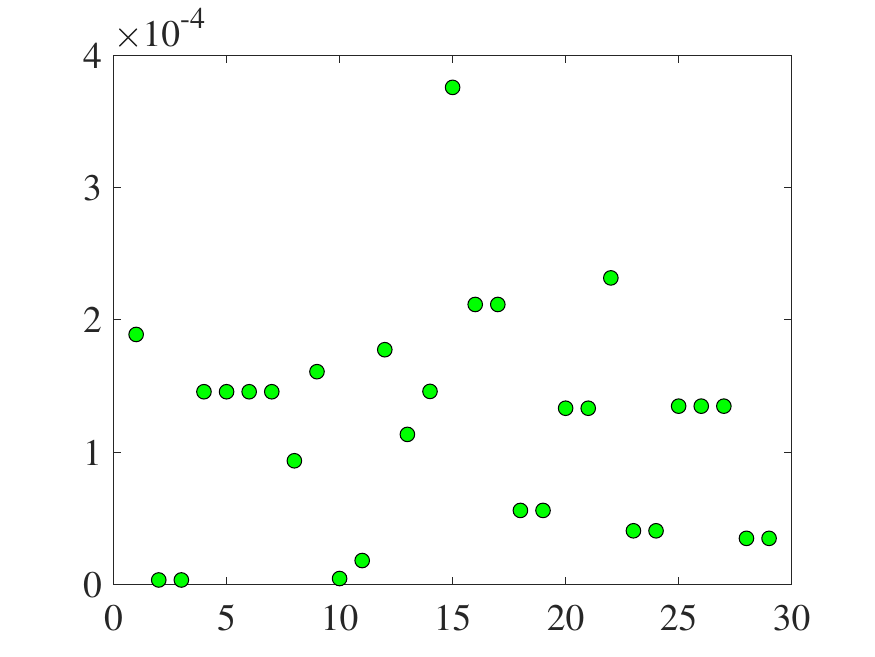}};
\node [anchor=west, font=\small] at (-0.90,-2.8) {\begin{tabular}{c} Time Step \end{tabular}};
\node [anchor=west, rotate=90, font=\small] at (-3.25,-2.2) { \begin{tabular}{c} Voltage Difference in Per-unit\end{tabular}};
\end{tikzpicture}%
\vspace{-.4cm}
 
	\caption{Voltage difference (in per-unit) between predicted and actual load at a representative bus near Parking Lot 6, showing that deviations remain small and validating the accuracy of the proposed forecasting model.}
	\label{fig:voltage_difference}
	   \vspace{-0.5cm}
\end{figure}

\rn{Fig.~\ref{fig:voltage_difference} shows the difference in bus voltage (in per-unit) between the predicted load and the actual load for one bus located near Parking Lot G6 in our model. As shown, the voltage difference is very small, indicating that the proposed algorithm accurately predicts the load. This accuracy is not only evident in the load estimation results but is also reflected in the physical quantities: the bus voltage and line power flow for the predicted load closely match the corresponding values for the actual load.}

\rn{The per-unit voltage deviations at representative buses across the campus under the two scenarios remain small, with maximum differences staying below 0.04\%. The largest variations occur at buses electrically close to major EV clusters, such as Lots G6 and B2. These deviations are most noticeable during afternoon peak charging hours, when actual loads experience sharp ramps. Even in these cases, the predicted voltage profiles capture both the magnitude and timing of these events closely, with deviations limited to approximately $4\times10^{-4}$ pu. This demonstrates that the forecasting model not only anticipates aggregate demand trends but also reflects their localized influence on voltage stability.}

% \begin{itemize}
%     \item \textbf{Limited Data:} Limited data points from a single year limit how well the model can generalize over multiple seasons or years. As a results the prediction windows that perform the best are limited to 30 day spans.
    
%     \item \textbf{No hyperparameter tuning:} The current design does not use an optimizer for hyperparameter tuning. This is due to the vast amount of time required for such training sessions. With a proper hyperparameter optimizer like PSO, one could see better accuracy.
% \end{itemize}

\section{CONCLUSION}
In this paper, an electric grid's model of California State University, Northridge, as a representative dense community, is found to understand and solve the complexity of EV charging infrastructure. AI-driven solutions is proposed to optimize electric grid efficiency for EV charging stations, reducing the risk of grid overload in dense urban areas. Additionally, it provides actionable insights for strategic energy planning, supporting the sustainable expansion of EV infrastructure. This line of research ultimately aims to create scalable, equitable solutions that enhance grid reliability, reduce energy disparities, and support the transition to renewable energy sources. Using innovative AI techniques, the proposed research will pave the way for smarter, more adaptive energy systems that can handle future demands efficiently and ultimately advance climate change solutions.
\section{ACKNOWLEDGMENT}
The authors gratefully acknowledge financial support for this research by a grant from the STEM-NET consortium at the CSU Office of the Chancellor. 
\bibliographystyle{ieeetr}
\bibliography{reference}
\addtolength{\textheight}{-3cm}

\end{document}